\documentclass[journal]{IEEEtran}
\usepackage{cite}

\ifCLASSINFOpdf
   \usepackage[pdftex]{graphicx}
   \graphicspath{{../pdf/}{../jpeg/}}
   \DeclareGraphicsExtensions{.pdf,.jpeg,.png}
\else
   \usepackage[dvips]{graphicx}
   \graphicspath{{../eps/}}
   \DeclareGraphicsExtensions{.eps}
\fi
\usepackage{amsmath}
\usepackage{array}

\ifCLASSOPTIONcompsoc
 \usepackage[caption=false,font=normalsize,labelfont=sf,textfont=sf]{subfig}
\else
 \usepackage[caption=false,font=footnotesize]{subfig}
\fi
\begin{document}
%
\title{FPCNet: Fast Pavement Crack Detection Network Based on Encoder-Decoder Architecture}
%
%
%

\author{Wenjun~Liu,
        Yuchun~Huang,
        Ying~Li,
        and~Qi~Chen

\thanks{W. Liu, Y. Huang and Q. Chen are with the School of Remote Sensing and Information Engineering, Wuhan University, Wuhan 430079, China (e-mail:  liuwenjun@whu.edu.cn; hycwhu@whu.edu.cn; chenqi1126@whu.edu.cn)}
\thanks{Y. Li is with the Mobile Sensing and Geodata Science Laboratory, Department of Geography and Environmental Management, University of Waterloo, ON N2L 3G1, Canada (e-mail: y2424li@uwaterloo.ca)}
}

%
%

\markboth{IEEE TRANSACTIONS ON INTELLIGENT TRANSPORTATION SYSTEMS, UNDER REVIEW.}%
{Shell \MakeLowercase{\textit{et al.}}: Bare Demo of IEEEtran.cls for IEEE Journals}
%



\maketitle

\begin{abstract}
Timely, accurate and automatic detection of pavement cracks is necessary for making cost-effective decisions concerning road maintenance. Conventional crack detection algorithms focus on the design of single or multiple crack features and classifiers. However, complicated topological structures, varying degrees of damage and oil stains make the design of crack features difficult. In addition, the contextual information around a crack is not investigated extensively in the design process. Accordingly, these design features have limited discriminative adaptability and cannot fuse effectively with the classifiers. To solve these problems, this paper proposes a deep learning network for pavement crack detection. Using the Encoder-Decoder structure, crack characteristics with multiple contexts are automatically learned, and end-to-end crack detection is achieved. Specifically, we first propose the Multi-Dilation (MD) module, which can synthesize the crack features of multiple context sizes via dilated convolution with multiple rates. The crack MD features obtained in this module can describe cracks of different widths and topologies. Next, we propose the SE-Upsampling (SEU) module, which uses the Squeeze-and-Excitation learning operation to optimize the MD features. Finally, the above two modules are integrated to develop the fast crack detection network, namely, FPCNet. This network continuously optimizes the MD features step-by-step to realize fast pixel-level crack detection. Experiments are conducted on challenging public CFD datasets and G45 crack datasets involving various crack types under different shooting conditions. The distinct performance and speed improvements over all the datasets demonstrate that the proposed method outperforms other state-of-the-art crack detection methods.
\end{abstract}


%
\IEEEpeerreviewmaketitle

\section{Introduction}
%
%
%
%
\IEEEPARstart{P}{avement} cracks, which are one of the most representative defects of roads, are mainly caused by overloading, temperature changes and road surface aging. These damages can degrade the performance of road surfaces, shorten the service life of roads, and endanger the driving safety of vehicles. Fast and accurate pavement crack detection facilitates timely maintenance of roads and prevents the road conditions from deteriorating further. With the rapid advancement in sensors and information technology (IT), millions of road images have been collected by many transportation agencies for crack detection.

Various manually designed features, such as grayscale \cite{1,2,3,4}, edge \cite{5,6,7}, Gabor filters \cite{8,9}, wavelet \cite{10,11}, and histogram of oriented gradients (HOG) \cite{12}, are used to detect cracks from images. However, owing to the complex and diverse topology, arbitrary shapes and varying widths, and the presence of oil spots, gravel, zebra crossings and other strong disturbances on roads that pose challenges to the identification and detection of cracks, the performance of these methods is still limited. In addition, the poor contrast around the cracked pixels caused by undesired imaging conditions (such as overexposure or underexposure) also makes crack detection difficult. Therefore, in complex situations, manually designing one or multiple robust features is ineffective for extracting cracks from different road images.

In deep learning, Convolutional Neural Networks (CNN) can automatically learn the characteristics of target objects through alternating layers of convolution and pooling, and subsequently classify them. Human experience for feature and classifier design is not required in such networks, which provides new opportunities for end-to-end crack detection. The CNN-based crack detection algorithms proposed by some researchers have achieved relatively successful results by training automatic crack feature learners. Some of these algorithms use object detection methods \cite{13}, \cite{14} or image block classification \cite{15}, \cite{16} to detect cracks. These algorithms can locate cracks in a pavement image but fail to detect them pixel by pixel. Some algorithms \cite{17}, \cite{18} first partition the crack image according to a certain size, and then predict whether single or multiple pixel(s) in the center of the block are cracks. Pixel-level prediction is achieved in these methods, but these methods are time consuming and do not involve end-to-end aspects. Some studies \cite{19}, \cite{20} applied a fully convolutional network (FCN) \cite{21} to crack detection to solve the above problems with high precision and speed. However, the FCN methods still have the following problems with respect to crack detection.

\begin{itemize}
\setlength{\itemsep}{0pt}
\setlength{\parsep}{0pt}
\setlength{\parskip}{0pt}
\item[1)] Pavement cracks have different widths and topologies, but the filters of the FCN methods use only one receptive field size to extract the crack features within only one context, thus limiting their robustness for crack detection

\item[2)] The edge, pattern or shape features of cracks contribute differently to the detection results. However, FCN methods treat these features equally with addition \cite{21} or concatenation \cite{22} operations performed during the lateral connection of different features.
\end{itemize}

Inspired by the Encoder-Decoder structure, we propose a new crack detection network called FPCNet.

First, the features of cracks of one context size are extracted through a series of convolutions and pooling layers in the Encoder.

A Multi-Dilation (MD) module is then used to obtain the crack features of multiple context sizes. The dilated convolution \cite{23} can increase the context and learn deeper features without compromising the edge resolution. It is employed in the MD module with multiple rates to extract the crack MD features of multiple context sizes. With the proposed MD module, the cracks of different widths and topologies can be robustly detected.

Next, an SE-Upsampling (SEU) module is developed to construct the Decoder. It restores the resolution of the crack MD features through transposed convolution. The features in the Encoder are added to the restored MD features of the same resolution, which combines the context of the cracks with edge details. After this addition, the SEU module assigns different weights to the MD features adaptively through the Squeeze-and-Excitation learning operation \cite{24}. The crack information corresponding to the edge, pattern or shape embedded in the MD features is assigned different weights, based on its contribution to the detection results.

Finally, the MD and SEU modules are integrated in the Encoder-Decoder structure to develop the fast crack detection network FPCNet. The network uniquely characterizes the crack context by using the MD module, and continuously optimizes the contextual features by using the SEU modules to obtain the finest prediction. Pixel-by-pixel crack detection is realized using the proposed FPCNet.

We conducted several experiments on public CFD datasets and G45 crack datasets involving multiple crack types under different shooting conditions. The experimental results show that FPCNet can detect multiple types of cracks and attain state-of-the-art precision on CFD datasets, with a high speed of 14.7 FPS.

The rest of the paper is organized as follows: Section II provides a brief review of crack detection methods; in Section III, we describe in detail our Multi-Dilation module, SE-Upsampling module and FPCNet; Section IV describes the performed series of experiments, and the corresponding results and analysis; Finally, Section V summarizes the main work presented in this article.

\section{Related Work}

Existing visual-based crack detection methods can be roughly classified into three categories: traditional, machine learning-based and deep learning-based methods. In this section, we briefly describe the application of these methods for crack detection.

\subsection{Traditional Method}

Early studies such as \cite{1,2,3,4} observed that the cracks in a road image are darker than the background; thus, different thresholding methods were used to extract the cracks. However, these methods experience difficulty in selecting the appropriate threshold. In addition, they are sensitive to imaging conditions and noise, which finally result in poor performance. Edge detection methods \cite{5,6,7} achieved distinct improvements in images with a large contrast between the crack edge and the background. However, these methods demonstrate limited performance in road images with low contrast or noise. The use of manually designed feature descriptors such as Gabor filters \cite{8}, \cite{9}, wavelet transform \cite{10}, \cite{11}, and histogram of oriented gradients (HOG) \cite{12} exhibit significant advancements in detecting simple cracks but they are not suitable for complex and diverse cracks. In addition, the parameter selection is commonly time-consuming and laborious.

\subsection{Machine Learning}

With the advancement of machine learning, the following methods have been successfully applied in crack detection: \cite{25} considered the road surface as a textured surface to design features, and then applied the support vector machine (SVM) for classification; \cite{26} utilized numerous linear and nonlinear filters to extract texture features that could then be filtered by AdaBoost; \cite{27} selected the random forest method to classify multiple spatially adjusted visual features. However, these detection methods are restricted to detecting learned cracks and find it difficult to detect new cracks. CrackForest \cite{28} solved this problem by using random structured forests classifiers, which can identify arbitrarily complex cracks. However, these methods are limited in terms of the quality and quantity of the manually designed features. Moreover, it is difficult to design universal features that can be applied to all types of cracks.

\subsection{Deep Learning}

Recently, deep learning has made great progress in the field of computer vision. The precision achieved by CNN-based networks has greatly exceeded the precision attained by traditional image classification methods \cite{29,30,31,32}, and even that possible at the human level \cite{32}. In recent research, deep learning-based methods have been successfully applied to road crack detection. \cite{14}, \cite{15} applied deep learning-based object detection methods to detect the location of cracks in road images; \cite{16}, \cite{17} utilized road grids or sliding windows to divide the road images into smaller image blocks before using a CNN to determine whether an image block contains a crack. Although the abovementioned methods can accurately locate the crack, they cannot detect cracks pixel by pixel. \cite{18}, \cite{19} selected image blocks in the road image via CNN to determine whether the central pixel or the pixels of the image block belong to the crack, which not only achieved pixel-by-pixel detection but also attained high precision. However, the small blocks fail to provide enough context information for prediction. Moreover, the time consumption is relatively high for block-based detection. \cite{20}, \cite{21} used the FCN network for crack detection and achieved high precision and speed. However, this method does not consider the fact that cracks with different widths and topologies require different context sizes. Moreover, in this method, the fact different crack features contribute differently to crack detection was ignored and all crack features were treated in the same manner.

\section{Methodology}

In this section, we first introduce the proposed Multi-Dilation module and SE-Upsampling module. Next, the network structure for crack detection, i.e., FPCNet is described.

\subsection{Multi-Dilation (MD)}

To extract crack features, the Encoder operation is performed first. This process includes four groups of two typical $3 \times 3$ convolutions. Each convolutional group produces its crack multiple-convolution (MC) features, which are then downsampled by a max pooling layer to capture the context. However, the convolutional filters in the Encoder use only one receptive field size to extract the crack features within one context. As a result, the MC features, which are extracted by the Encoder, cannot robustly detect cracks with different widths and topologies.

Thus, the Multi-Dilation (MD) module is developed, which is based on the MC features. The dilated convolution, which expands the context window size of the convolution without downsampling or convolving with larger filters of more parameters, is employed in the MD module. By combining multiple dilated convolutions \cite{23} with different rates and a global pooling, the MD module extracts crack features with multiple context sizes and detects cracks with different widths and topologies.

The dilated convolution was first proposed by \cite{23} to efficiently perform wavelet decomposition. In a 1D signal, the dilated convolution can be defined as follows:

\begin{equation}
\label{equa_1}
    y[i] = \sum_{k=1}^{K}x[i+r\cdot k]w[k]
\end{equation}where $x[i]$ is the input signal, $y[i]$ is the output signal, $w[k]$ represents the filter of length $K$, and parameter $r$ represents the interval at which the dilation is used to sample the input signal. In default filtering operation, $r=1$.

The dilated convolution is used in convolution operations \cite{33, 34}, which add ``holes" with a value of 0 between the pixels of the convolution kernel. For a convolution kernel of size $k\times k$, the actual convolution kernel size $k_{result} = k + (k-1) \times (r-1)$. As shown in Fig. 1, (a) is the default convolution, and (b), (c) and (d) are the convolution kernels for kernels for $r=2$, $3$, and $4$, respectively.

\begin{figure}[!t]
\centering

\subfloat[$r=1$]{\includegraphics[width=0.8in]{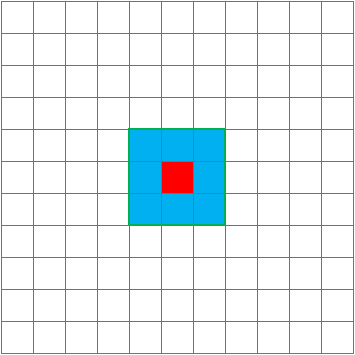}%
\label{r=1}}
\hfil
\subfloat[$r=2$]{\includegraphics[width=0.8in]{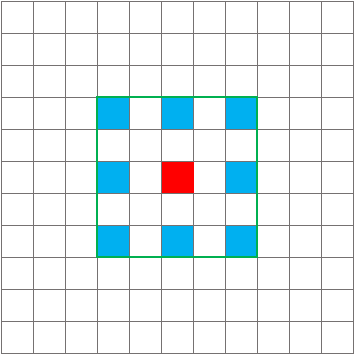}%
\label{r=2}}
\hfil
\subfloat[$r=3$]{\includegraphics[width=0.8in]{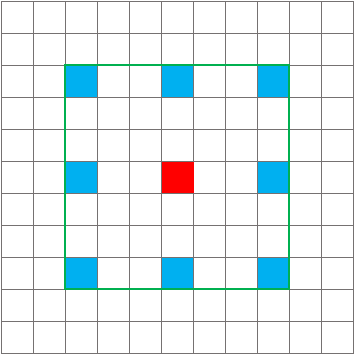}%
\label{r=3}}
\hfil
\subfloat[$r=4$]{\includegraphics[width=0.8in]{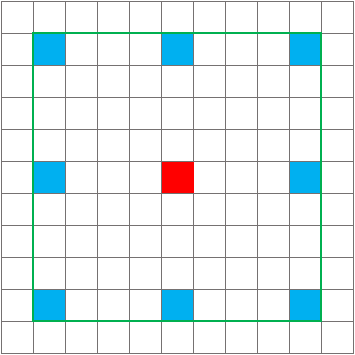}%
\label{r=4}}
\hfil

\centering
\caption{Convolution kernels with different dilation rates.}
\label{fig_1}
\end{figure}

As shown in Fig. \ref{fig_1}, the context of the dilated convolution kernel is larger than that of the standard convolution kernel when $r$ is greater than 1. Since ``0" is not a parameter, the parameters and calculation amount of the convolution kernel are not actually increased. Thus, compared to the standard convolution with larger filters, dilated convolution enlarges the context of the convolution operation without increasing the amount of calculation involved.

However, in complex road images, the width along the same crack curve changes dramatically. In addition, the contexts required for detecting cracks of different topologies and severity levels are different, and dilated convolution with one rate can only get one context. For example, when $r = 1$, which is the standard convolution, the context size obtained is small. Such a convolution is suitable for thin and simple cracks, but it cannot effectively detect wide cracks as well as cracks with complex topologies. However, these cracks can be robustly detected by dilated convolutions with a larger value of $r$ (for example, 4). Thus, a dilated convolution with a single rate cannot obtain all the required contextual information for crack detection.

\begin{figure}[!t]
\centering
\includegraphics[width=3.5in]{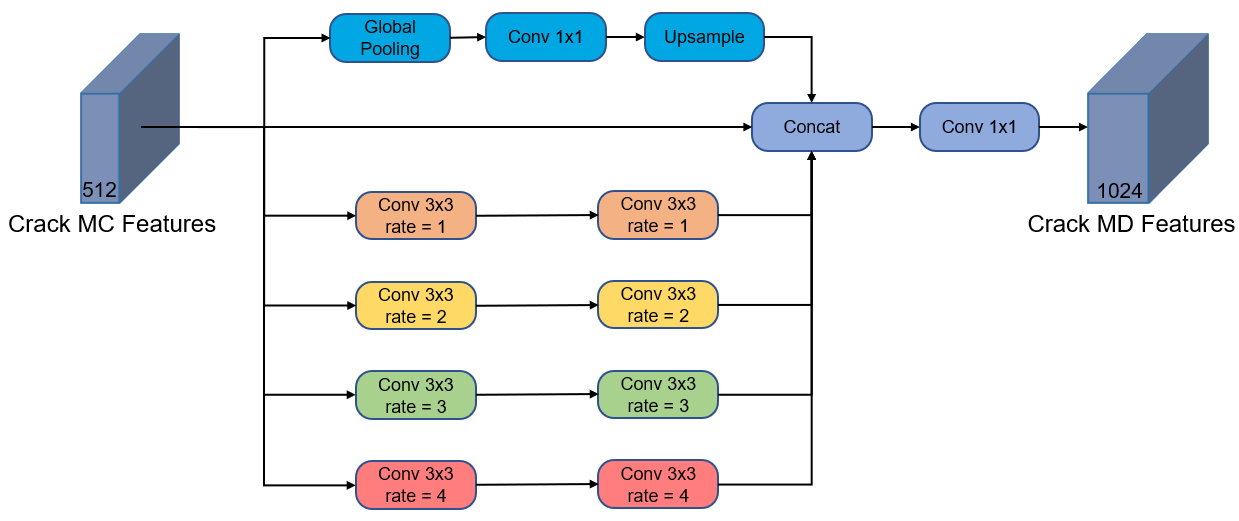}

\centering
\caption{Multi-Dilation module. The module concatenates four dilated convolutions with rates of \{1, 2, 3, 4\}, a global pooling layer and the original crack MC features. After the concatenation, a $1\times 1$ convolution is performed to obtain the crack MD features. Every convolution retains its number of feature channels except the last $1\times 1$ convolution, and padding is used to ensure that the resolution of the MC feature remains constant.}
\label{fig_2}
\end{figure}

Based on this, we propose the Multi-Dilation module. As shown in Fig. \ref{fig_2}, the input is the crack MC features extracted from the Encoder, and the output is the crack MD features. This module analyzes crack features with different context sizes and integrates them to obtain features with multiple contexts. First, the dilated convolutions with four rates \{1, 2, 3, 4\} are used to obtain the crack features of different context sizes. The global pooling layer is then added to obtain the global crack information contained in the MC features. To retain the crack information of the MC features, they are fed directly into the final output. Next, a concatenation method is employed to combine the abovementioned six features from pixelwise to global context. Subsequently, the number of feature channels is increased to six times that of the original number of channels, which improves the amount of calculation in subsequent operations. As a result, a $1\times 1$ convolution is finally applied to reduce the number of channels from $512\times 6$ for the concatenated features to 1024, which also increase the communication among these channels. After performing this convolution, we can obtain the output, that is, the crack MD features.

\begin{figure*}[t]
\centering
\includegraphics[width=7.0in]{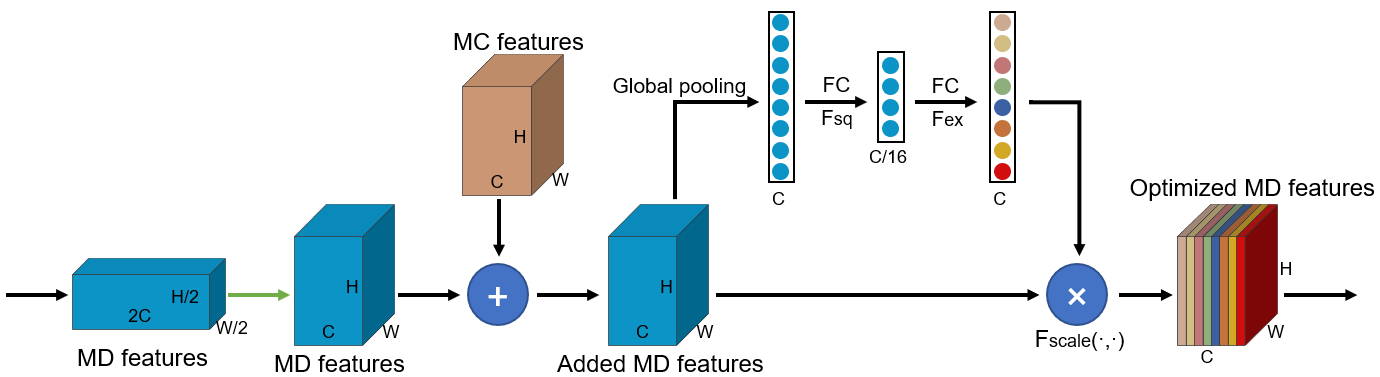}

\centering
\caption{SE-Upsampling module. The MC features are first added to the MD features after transposed convolution. Next, global pooling is performed to obtain the global information of the $C$ channels. After squeeze ($F_{sq}$) and excitation ($F_{ex}$) (two fully connected layers) of the global information, the weight of each feature for its channel is obtained. Finally, each feature in the added MD features is multiplied ($F_{scale}$) by its corresponding weight to obtain the optimized MD features. The green arrow indicates the transposed convolution. $H$, $W$, and $C$ represent the length, width, and number of channels of the features, respectively.}
\label{fig_3}
\end{figure*}

Note that the MD module integrates the features of multiple context sizes, including the pixelwise context, contexts of multiple dilated convolution rates and global context, which can help robustly describe cracks of various widths and topologies. The rates of the dilated convolutions in the MD module are set according to the statistics of the crack widths, and can be readily expanded, if necessary, for different cases.

\subsection{SE-Upsampling (SEU)}

The resolution of the crack MD features decreases because it is based on the MC features of many subsamplings. To achieve pixel-level detection, the resolution of the MD features needs to be restored to that of the original input pavement image. Hence, the Decoder operation is performed. With the Decoder’s upsampling operation (such as transposed convolution or bilinear interpolation), the resolution of the MD features can be continuously restored. Owing to the involvement of fewer subsamplings compared to those of the MD features, the MC features in the Encoder have more crack details, which are blurred in the MD features. To incorporate more crack details, the MD features can be combined with the MC features of the same resolution, i.e., lateral connection of the MD and MC features can be realized. However, if two different features are simply concatenated or added as proposed in \cite{21,22}, the different contributions of the edge, pattern, texture and other information embedded in the features for crack detection are regarded as being identical.

To overcome this problem, we propose the SE-Upsampling (SEU) module, as shown in Fig. \ref{fig_3}. The inputs are MD features and MC features, and the output is the optimized MD features after weighted fusion. The SEU module first restores the resolution of the crack MD features through transposed convolution. Next, it adds the MC features to the MD features in order to fuse the associated crack information concerning the edge, pattern, texture among others. Subsequently, the Squeeze-and-Excitation learning operation \cite{24} is applied to the added MD features to learn the weights of the different features. After the learning, the SEU module can adaptively assign different weights to different crack features such as the edge, pattern, and texture.

Specifically, the MD features first undergo upsampling by transposed convolution, which restores their resolution by 2 times and reduces their number of channels to half of the original value. Next, the MC features with the same resolution are added to the MD features. Global average pooling is performed to obtain the global information of each channel from the added MD features. Subsequently, the global information is processed by a squeeze operation ($F_{sq}$). A fully connected layer is used to squeeze the number of channels with a certain ratio (in this study, we use a ratio of $\frac{1}{16}$) and the ReLU layer is used to nonlinearize the output. We carry out an excitation process ($F_{ex}$) on the output, which restores the squeezed output to its original number of channels by using a fully connected layer. The sigmoid layer is used to obtain the channel weights. A larger weight indicates that the feature in a channel has a larger contribution to crack detection. Finally, each MD feature is multiplied ($F_{scale}$) by its corresponding weight to obtain the optimized MD feature.

\subsection{Network Architecture}

\begin{figure}[t]
\centering
\includegraphics[width=3.5in]{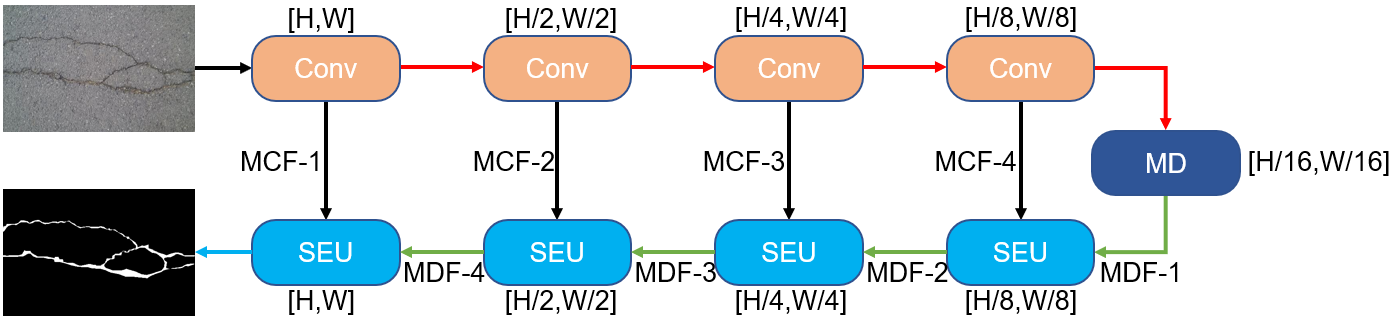}

\centering
\caption{Network architecture of FPCNet. The method uses 4 Convs (two $3\times 3$ convolutions and ReLUs) + max poolings as the Encoder to extract features. Next, the MD module is employed to obtain the information of multiple context sizes. Subsequently, 4 SEU modules are operated as the Decoder. $H$ and $W$ indicate the original sizes of the image. The red, green, and blue arrows indicate the max pooling, transposed convolution and $1 \times 1$ convolution + sigmoid, respectively. MCF denotes the multiple-convolution features extracted in the Encoder, and MDF denotes the MD features.}
\label{fig_4}
\end{figure}

FPCNet is developed by integrating the Multi-Dilation module and the SE-Upsampling module in the Encoder-Decoder structure. The network structure of FPCNet is shown in Fig. \ref{fig_4}.

This framework shares the common Encoder-Decoder structure of semantic segmentation networks: The upper row is the Encoder structure, and the bottom row is the Decoder structure. Each Conv in the Encoder structure consists of two typical $3 \times 3$ convolutional layers followed by a nonlinearization layer ReLU. Each $3 \times 3$ convolution is padded to maintain the original resolution. After the convolution, the $2 \times 2$ max pooling layer is used for downsampling. In total, there are four sets of convolutions + pooling operations. After each operation, the resolution of the MC features is reduced to half of their original resolution, and the number of channels is increased by two times, that is, the number of channels in each group is \{64, 128, 256, 512\}.

After the fourth max pooling layer at the end of the Encoder, the proposed Multi-Dilation module is employed to extract the contextual crack MD features of multiple sizes for robust crack detection.

The Decoder process is constructed by four successive SEU modules. The number of channels of the MD features are gradually reduced, specifically \{512, 256, 128, 64\}, and the resolution of the MD features is restored. By continuously operating the SEU modules, the MD features are also optimized step-by-step to obtain the finest prediction.

Finally, after the Decoder operation, the information in the feature vector of each pixel is integrated and predicted by a $1\times 1$ convolution, and later, the sigmoid nonlinearization layer is used to maintain the prediction probability between 0 and 1.

With the proposed FPCNet, fast pixel-level crack detection can be achieved owing to the following factors:
\subsubsection{MD module} Since the crack MC features undergo four iterations of max poolings, the size of the input MC features in the MD module is small (for example, after four runs of downsamplings, the size reduces from $512\times 512$ to $32\times 32$). As a result, although five crack contextual features with different context sizes are calculated from the MC features to obtain the MD features, the calculation cost does not increase sharply.

\subsubsection{SEU module} Instead of the concatenation operation described in \cite{22}, the SEU module uses the addition operation to combine two types of features, which reduces the subsequent amount of required calculation.

\subsubsection{Encoder-Decoder structure} FPCNet uses an end-to-end structure to achieve pixelwise prediction, which is much faster than the block-based methods.

\section{Experiments}

This section first describes the evaluation of the proposed approach on two crack datasets: public CFD \cite{28} and our own G45 crack dataset. Next, the selection of hyperparameters of MD module is discussed.

We have implemented our approach using Pytorch \cite{35} as the deep learning framework for training and evaluation under a PC with an operating system of Windows 10, which has an Intel(R) Core (TM) i7-6800K CPU @ 3.40 GHz with 16 GB memory and a NVIDIA GTX1080Ti GPU with 11 GB memory. To evaluate the proposed approach, we compare it with FCN \cite{21} and other state-of-the-art methods tested on the CFD dataset, including CrackForest \cite{28}, MFCD \cite{36}, method \cite{37}, and method \cite{18}. At the same time, to verify the effectiveness and scalability of FPCNet, the network is applied and evaluated on our G45 dataset including various type of cracks.

\emph{Evaluation:} To evaluate the performance of the proposed network, the values of Precision, Recall and F1 score are introduced. These values are computed based on true positives (TP), true negatives (TN), false positives (FP) and false negatives (FN) as follows:

\begin{equation}
    Precision = \frac{TP}{TP + FP}
\end{equation}

\begin{equation}
    Recall = \frac{TP}{TP + FN}
\end{equation}

\begin{equation}
    F1\quad score = \frac{2 \times Precision \times Recall}{Precision + Recall}
\end{equation}

Because both Precision and Recall have their biases, this study focuses on the F1 score. Since it is extremely difficult for the pixel-level ground truth to be obtained by crack images, a tolerance margin is used in most crack detection algorithms for evaluating the performance of the algorithm. This margin takes into account detected pixels that are no more than 2 \cite{18,37} or 5 \cite{28,36} pixels away from the ground truth as the TP. We use a tolerance margin of 2 pixels in this study.

\subsection{CFD Dataset}

The CFD dataset was published in \cite{28}, and it is composed of 118 RGB images with a resolution of $480\times 320$ pixels. All the images are taken using an iPhone 5 from pavements of Beijing, China, which can generally reflect the urban pavement surface condition existing in Beijing. These images have uneven illumination and contain noises such as shadows, oil spots and water stains, which make crack detection quite difficult. We randomly divided 60\% (72 images) of the dataset for training and 40\% (46 images) of the dataset for testing, as described in \cite{28}.

\subsubsection{Implementation details}An insufficient number of crack images can easily cause the problem of overfitting during training. Thus, data augmentation is performed, including clockwise rotation of $90^{\circ}$ and $180^{\circ}$, horizontal flip, and random color jittering. Random cropping to a size of $288\times 288$ is performed in training. Data augmentation is not used during testing.

We use the following binary cross entropy (BCE) + dice coefficient loss as the loss function during training:

\begin{equation}
\begin{split}
L(Y^{*},Y) = &\frac{1}{N}\sum_{P\in{N}}(Y^{*}_{P} \cdot \lg Y_{P} + (1 - Y^{*}_{P}) \cdot \lg(1- Y_{P}) \\
&+1 - \frac{2 \times TP}{2 \times TP + FP + FN}
\end{split}
\end{equation}where $Y^{*}$ and $Y$ denote the target image and prediction image, respectively; N is the set of all pixels in the image; and $Y_{P}$ and $Y_{P}^{*}$ denote the values at pixel $p$ in the prediction and the target images, respectively. We use the initialization method proposed in \cite{38} as the weight initialization approach, and choose SGD with Momentum (0.9) \cite{39} as our optimizer with a batch size of 1 and a weight decay of 0.0001. Training is started with a learning rate of 0.01; it is reduced by 10 at epochs 50, 80, and 110, and training is terminated at 120 epochs.

\begin{figure}[t]
\centering
\subfloat{\includegraphics[width=1.12in]{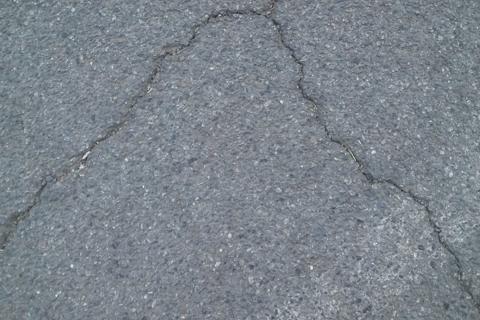}%
\label{5-1-1}}
\hfil
\subfloat{\includegraphics[width=1.12in]{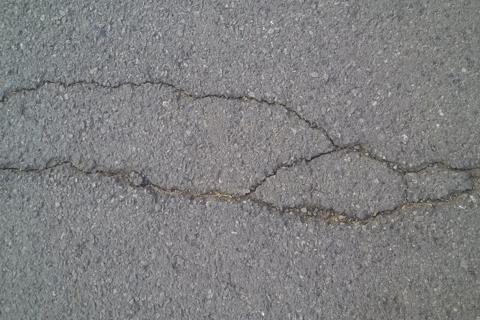}%
\label{5-1-2}}
\hfil
\subfloat{\includegraphics[width=1.12in]{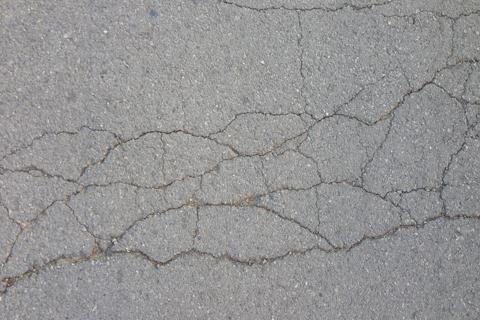}%
\label{5-1-3}}
\vspace{-5pt}

\subfloat{\includegraphics[width=1.12in]{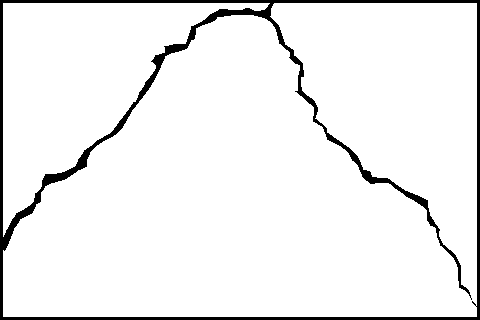}%
\label{5-2-1}}
\hfil
\subfloat{\includegraphics[width=1.12in]{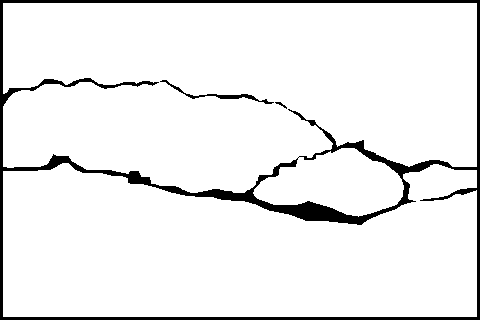}%
\label{5-2-2}}
\hfil
\subfloat{\includegraphics[width=1.12in]{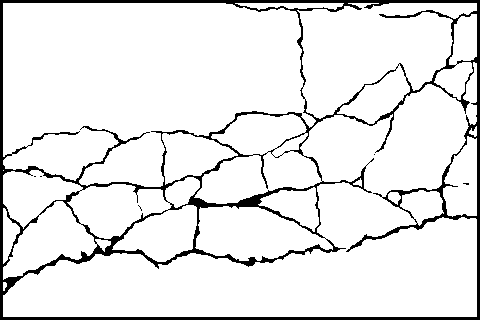}%
\label{5-2-3}}
\vspace{-5pt}

\subfloat{\includegraphics[width=1.12in]{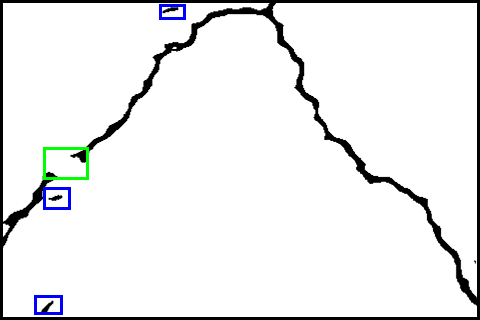}%
\label{5-3-1}}
\hfil
\subfloat{\includegraphics[width=1.12in]{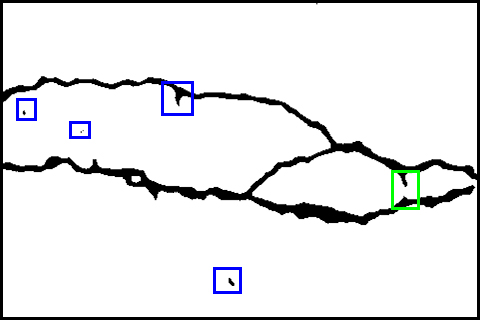}%
\label{5-3-2}}
\hfil
\subfloat{\includegraphics[width=1.12in]{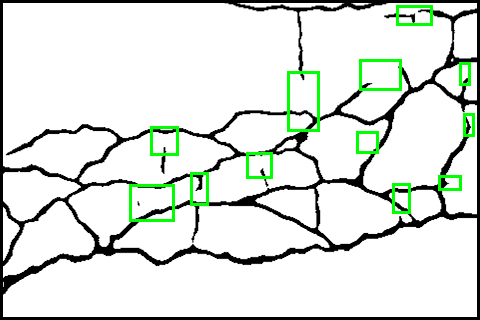}%
\label{5-3-3}}
\vspace{-5pt}

\subfloat{\includegraphics[width=1.12in]{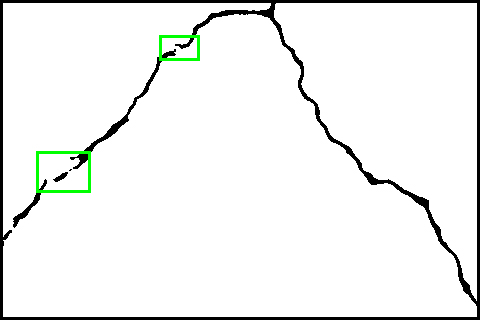}%
\label{5-4-1}}
\hfil
\subfloat{\includegraphics[width=1.12in]{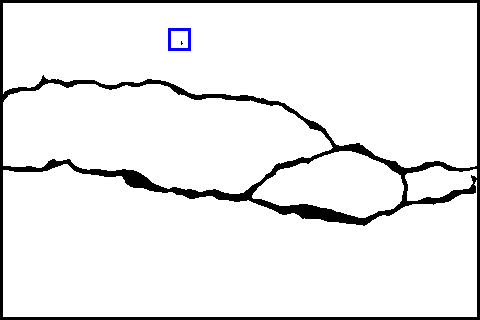}%
\label{5-4-2}}
\hfil
\subfloat{\includegraphics[width=1.12in]{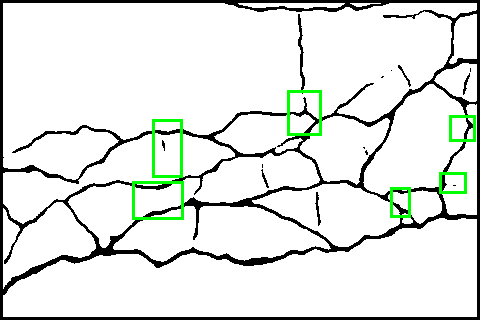}%
\label{5-4-3}}
\vspace{-5pt}

\subfloat{\includegraphics[width=1.12in]{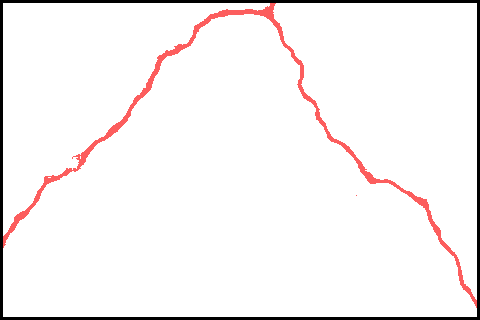}%
\label{5-5-1}}
\hfil
\subfloat{\includegraphics[width=1.12in]{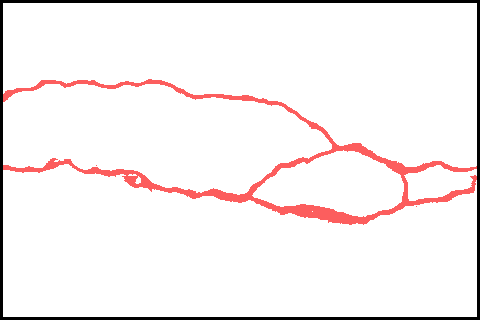}%
\label{5-5-2}}
\hfil
\subfloat{\includegraphics[width=1.12in]{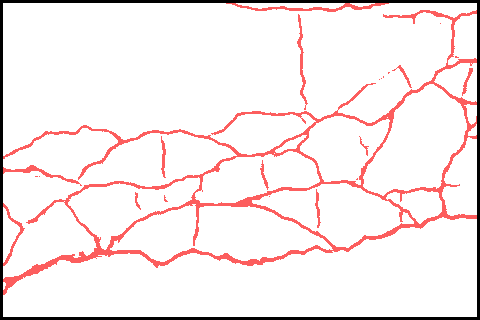}%
\label{5-5-3}}
\vspace{-5pt}

\subfloat{\includegraphics[width=1.12in]{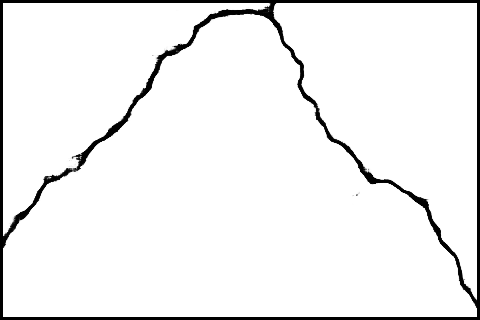}%
\label{5-6-1}}
\hfil
\subfloat{\includegraphics[width=1.12in]{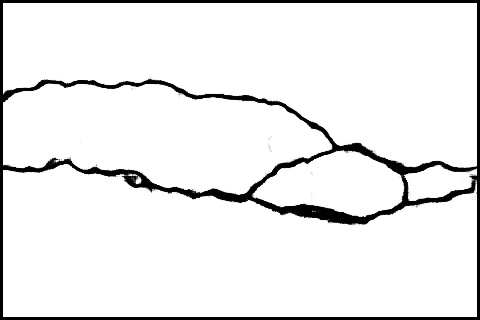}%
\label{5-6-2}}
\hfil
\subfloat{\includegraphics[width=1.12in]{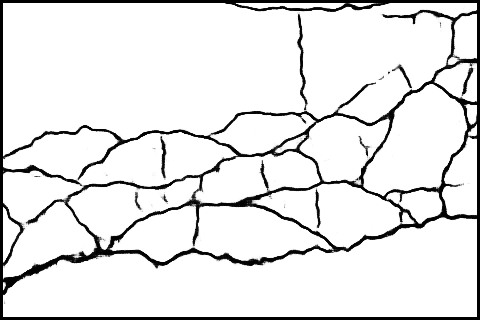}%
\label{5-6-3}}
\vspace{-5pt}

\subfloat{\includegraphics[width=1.12in]{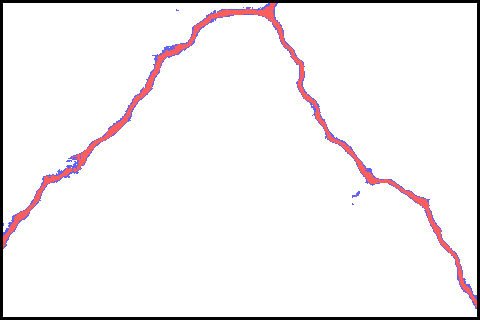}%
\label{5-7-1}}
\hfil
\subfloat{\includegraphics[width=1.12in]{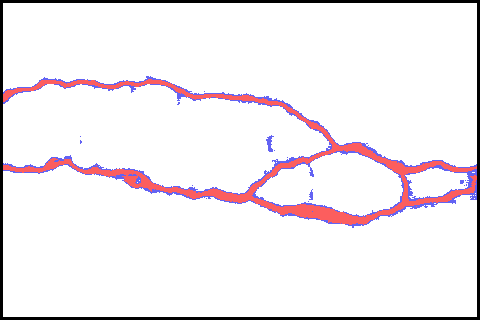}%
\label{5-7-2}}
\hfil
\subfloat{\includegraphics[width=1.12in]{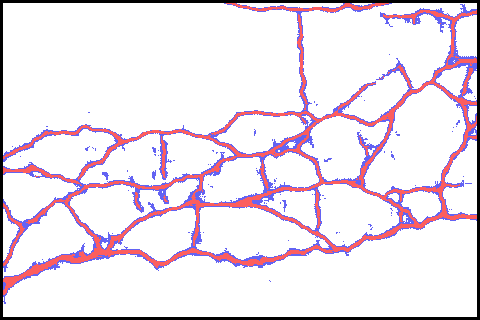}%
\label{5-7-3}}
\hfil
\centering
\caption{Results of comparison of proposed approach with Method \cite{18} on CFD (from top to bottom: original image, ground truth, Method \cite{18}, FCN \cite{21}, FPCNet, probability images predicted by FPCNet, special display of the probability images).}
\label{fig_5}
\end{figure}

\subsubsection{Results}We compare the proposed approach with FCN \cite{21} (We trained the network on the CFD dataset) and four other crack detection methods tested on the CFD dataset: CrackForest \cite{28}, MFCD \cite{36}, Method \cite{37}, and Method \cite{18}. The result shown in TABLE \ref{table_1} demonstrates that our approach outperforms the others. FPCNet achieves an F1 score of 96.93\%, thereby exceeding the F1 score of Method \cite{18} by 4.49\%, which is the current state-of-the-art approach on the CFD dataset and that of FCN by 1.03\%. Fig. \ref{fig_5} shows the comparison of the proposed approach with Method \cite{18} and FCN \cite{21}. In this figure, wrong detection (FP) and missed detection (TN) are indicated by blue and green rectangles, respectively. As seen from the third row in the figure, several wrong detections and missed detections occur in the images predicted by Method \cite{18} because of the method’s small context of the block for detection. The FCN with a large context size solves these problems to some extent. As shown in the fourth row, most of the noise (wrong detection) is eliminated and the situation of missed detections is also alleviated. However, owing to the FCN’s single context size and because it treats all features equally during detection, a large number of missed detections (green rectangles in the fourth row) occur, especially for cracks having complex topologies. Compared with these two methods, the results predicted by FPCNet have fewer wrong detections and missed detections, as seen in the fifth row of Fig. \ref{fig_5}, indicating the robustness of our approach. This is because FPCNet can acquire the features of multiple context sizes via the MD module and treat them differently according to their respective contributions via the SEU module.

The sixth row shows the probability images predicted by FPCNet, in which a darker pixel has a higher probability of being a crack. It can be observed that the noises and pixels near the outer sides of the crack edges have such a low prediction probability that they are barely noticeable. To illustrate these features clearly, we indicate the pixels with probabilities higher than 0.5 in red and those with probabilities lower than 0.5 in blue in the seventh row (all the following probability images are marked in a similar manner). After binarization, the noises and pixels on the outer sides, which have low probabilities (marked in blue), are erased. This indicates that, owing to the design of the MD module, FPCNet can obtain wide-ranging and multiple contextual information to effectively suppress noise and adapt to the different widths of the cracks.

\begin{table}[!t]
\renewcommand{\arraystretch}{1.5}
\caption{RESULTS FOR CRACK DETECTION EVALUATION ON CFD DATASET}
\label{table_1}
\centering
\begin{tabular}{ccccc}
\hline
Method & Tolerance Margin & Precision & Recall & F1 score\\
\hline
CrackForest\cite{28} & 5 & 82.28\% & 89.44\% & 85.71\% \\

MFCD\cite{36} & 5 & 89.90\% & 89.47\% & 88.04\%\\
Method\cite{37} & 2 & 90.70\% & 84.60\% & 87.00\%\\
Method\cite{18} & 2 & 91.19\% & 94.81\% & 92.44\%\\
FCN\cite{21} & 2 & 97.29\% & 94.56\% & 95.90\%\\
FPCNet & 2 & \textbf{97.48\%} & \textbf{96.39\%} & \textbf{96.93\%}\\
\hline
\end{tabular}
\end{table}

\subsubsection{Time comsumption} 
\begin{figure}[t]
\centering
\includegraphics[width=3.5in]{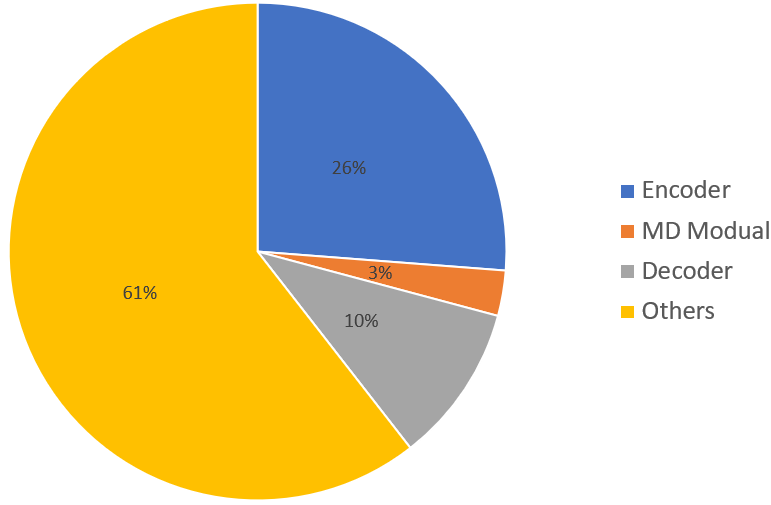}

\centering
\caption{Percentage of time consumption of each module in the process of crack detection by FPCNet in the CFD dataset. The total time consumption is 67.9 ms per image.}
\label{fig_6}
\end{figure}

The analysis of the time consumption of FPCNet is discussed herein. We first test the time required by each module in FPCNet during the detection of one image. The corresponding results are shown in Fig. \ref{fig_6}. It can be observed intuitively that the Encoder takes the most time (17.8 ms) in FPCNet. This is because the Encoder’s eight convolutional layers and four pooling layers are all operated for the large images of fine crack MC features. As for the MD module, even if the module contains 10 convolutional layers, one global pooling layer, and one upsampling layer, it is operated after the extraction of MC features. Consequently, the time consumption is only 2.0 ms due to the small sizes of the features. The Decoder module has 4 SE-Upsampling modules and an additional predicted convolution layer. Although the image resolution increases, the process takes only 7.0 ms, which is only 1/8th the time taken by the Encoder module. Other operations including image reading, processing, and predicted image saving, take up 41.09 ms. The total time consumption in predicting an image is 67.9 ms, which corresponds to a speed of 14.7 frames per second.

We also compare the time consumption with Method \cite{18} since it is the current state-of-the-art approach on the CFD dataset. We reproduce this method by Pytorch and test the average time consumption of one crack image with the same requirement of GPU memory (890 M) and environment (CPU: Intel(R) Core(TM) i7-6800K CPU @ 3.40 GHz, Memory: 16 GB, GPU: NVIDIA GTX1080Ti). The speed of the FPCNet for the detection of an image is 5.7 times that of Method \cite{18}, as shown in TABLE \ref{table_2}. This proves that our network demonstrates not only a higher detection accuracy but also faster detection speed, and it is suitable for large-scale pavement crack detection.

\begin{table}[!t]
\renewcommand{\arraystretch}{1.5}
\caption{RESULTS FOR TIME CONSUMPTION ON CFD DATASET}
\label{table_2}
\centering
\begin{tabular}{cccc}
\hline
Method & Batch Size & Time Comsumption & FPS\\
\hline
Method\cite{18} & 2800 & 380ms & 2.6\\
FPCNet & 1 & \textbf{67.9ms} & \textbf{14.7}\\
\hline
\end{tabular}
\end{table}

\subsection{G45 Dataset}

\begin{figure*}[p]
\centering

\subfloat{\includegraphics[width=1.65in]{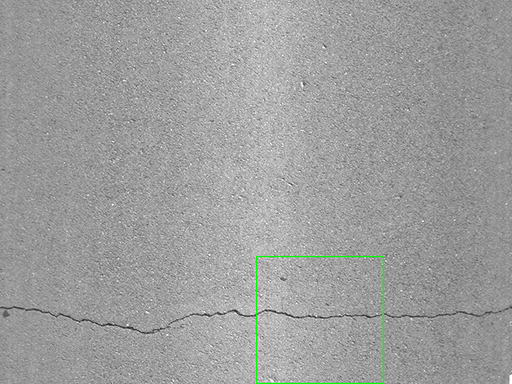}%
\label{7-1-1}}
\hfil
\subfloat{\includegraphics[width=1.65in]{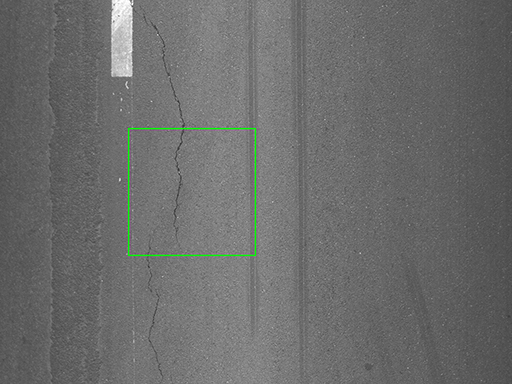}%
\label{7-1-2}}
\hfil
\subfloat{\includegraphics[width=1.65in]{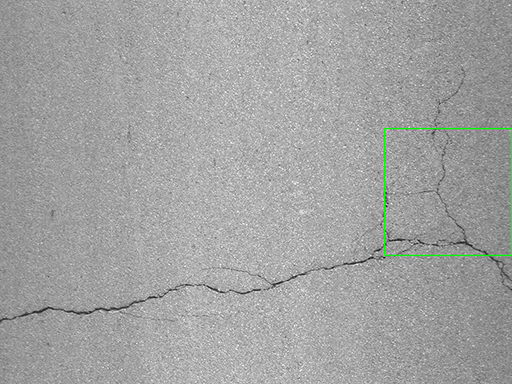}%
\label{7-1-3}}
\hfil
\subfloat{\includegraphics[width=1.65in]{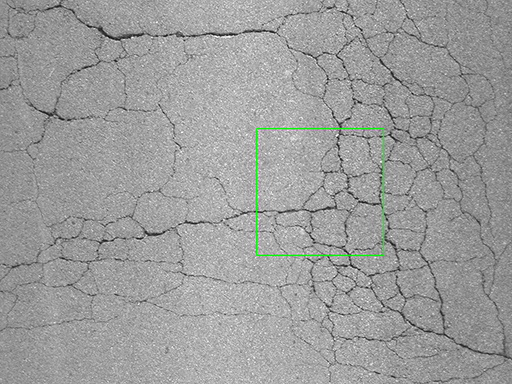}%
\label{7-1-4}}
\vspace{-5pt}

\subfloat{\includegraphics[width=1.65in]{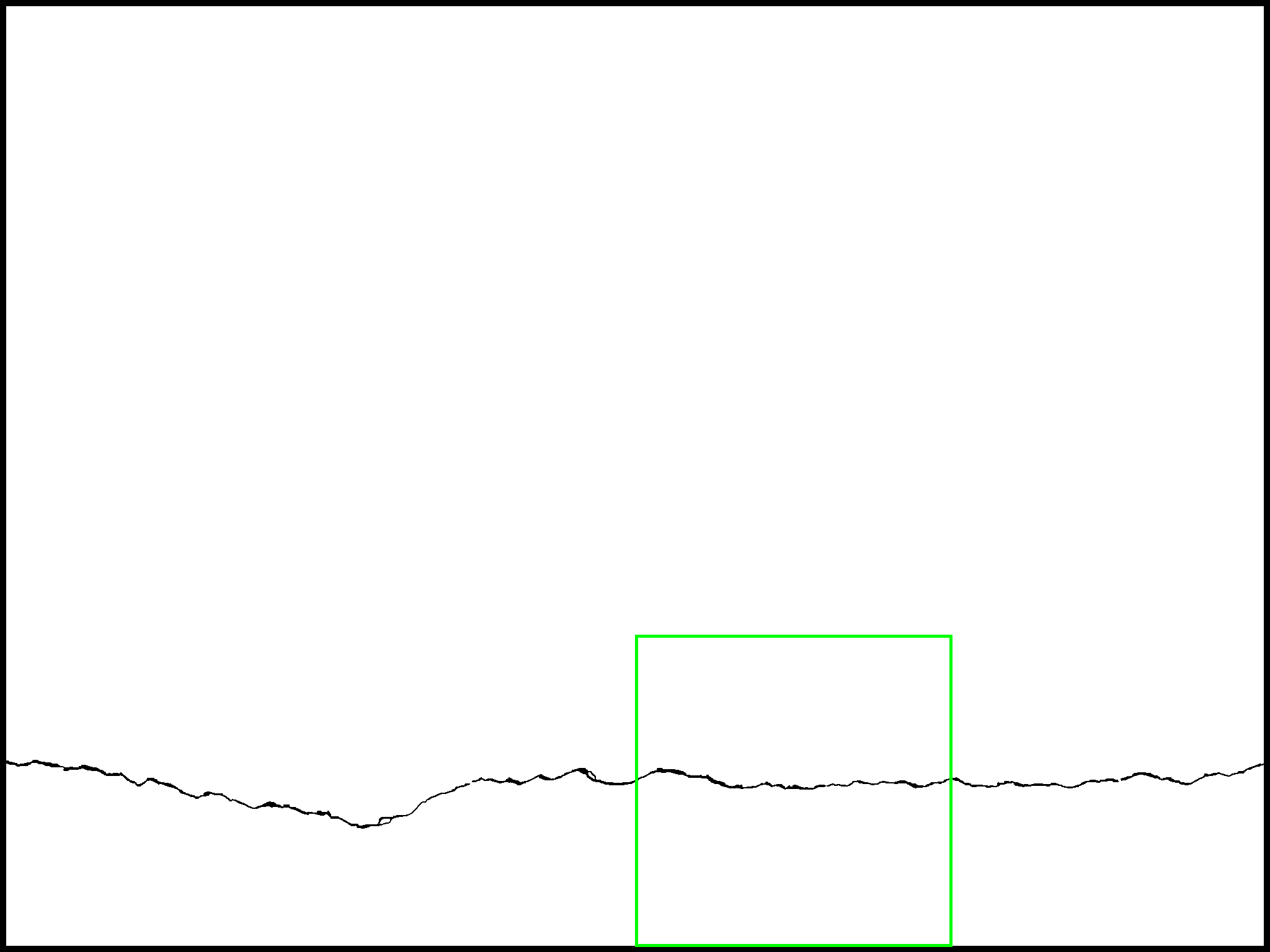}%
\label{7-2-1}}
\hfil
\subfloat{\includegraphics[width=1.65in]{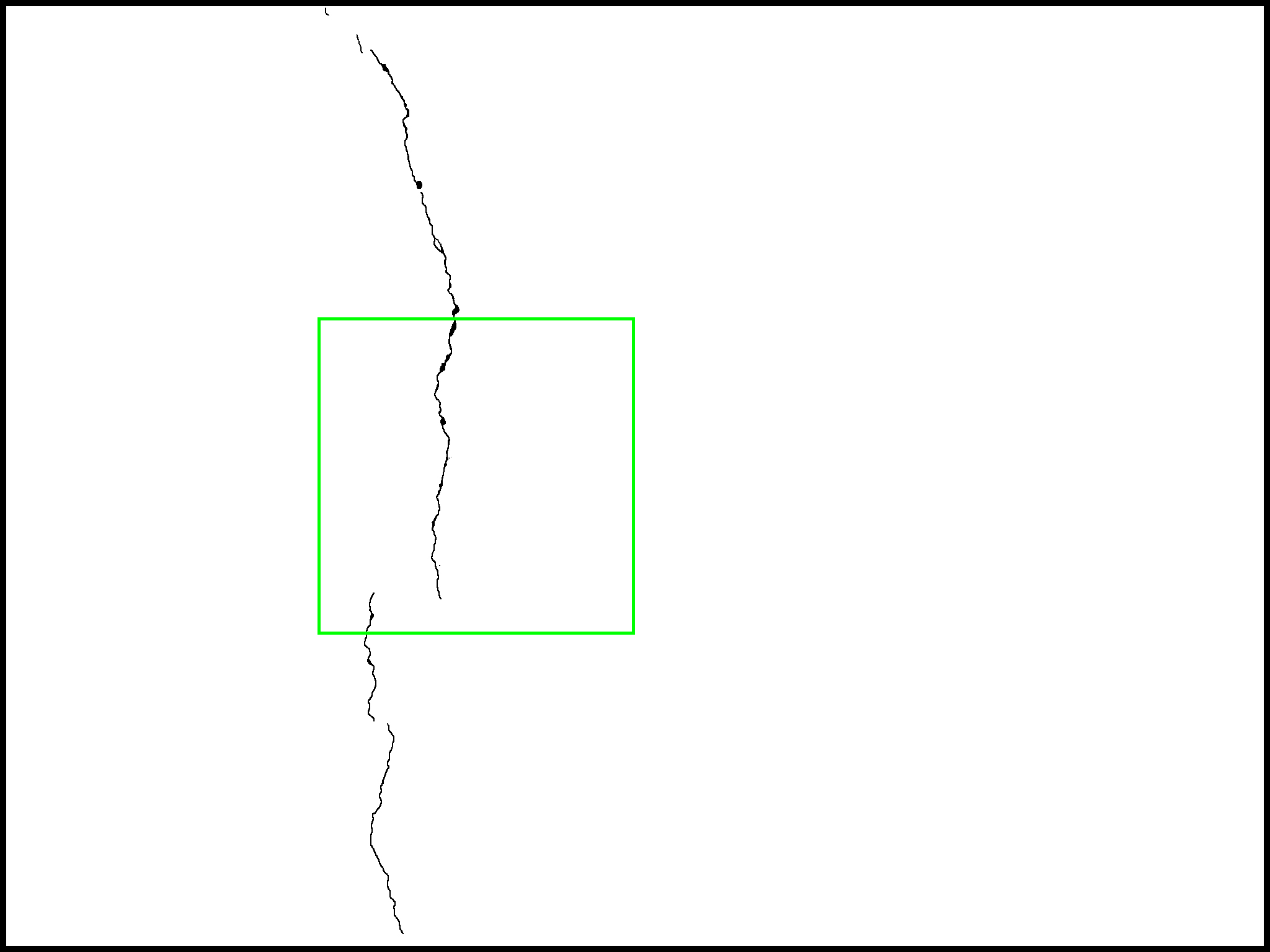}%
\label{7-2-2}}
\hfil
\subfloat{\includegraphics[width=1.65in]{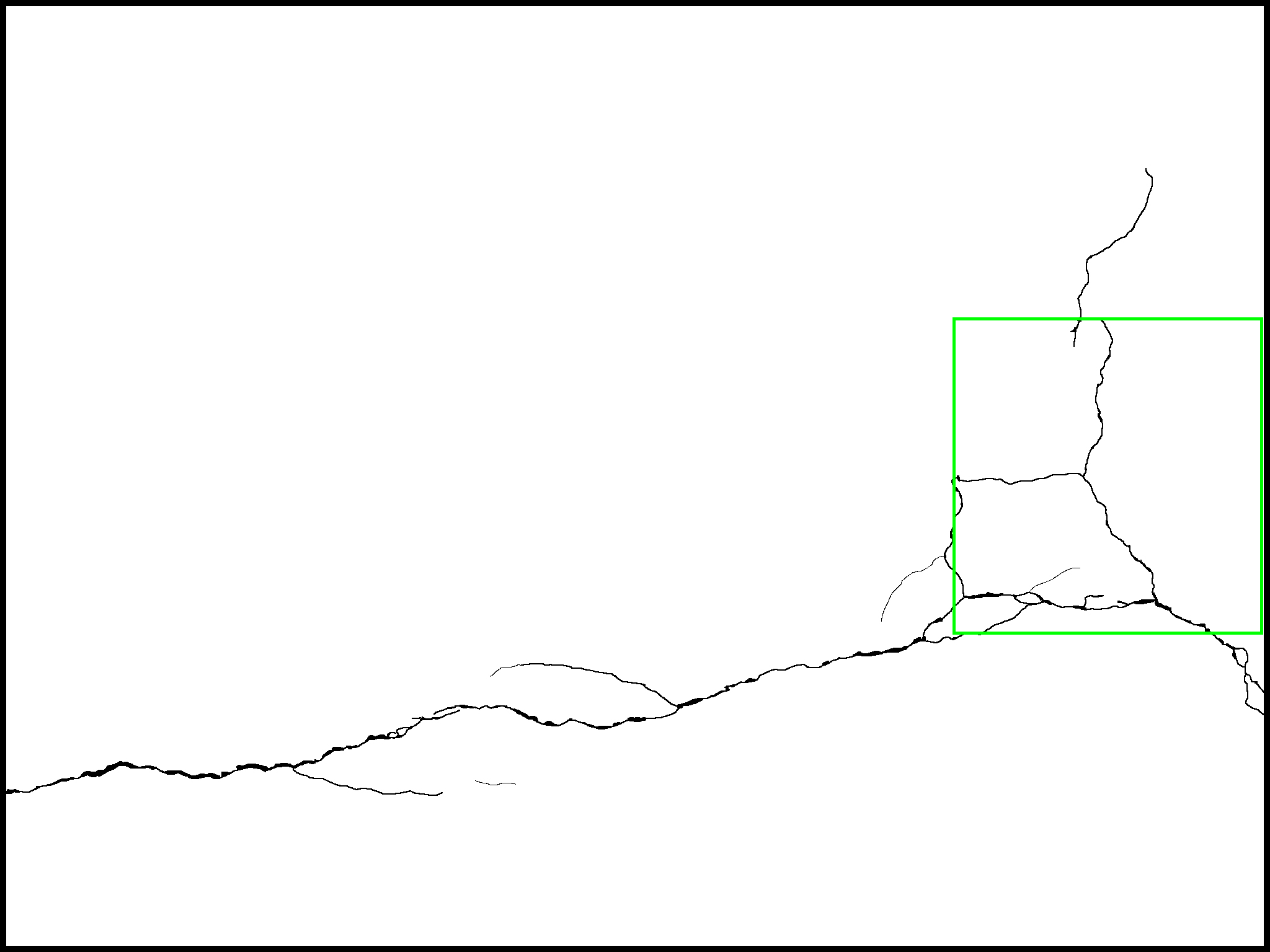}%
\label{7-2-3}}
\hfil
\subfloat{\includegraphics[width=1.65in]{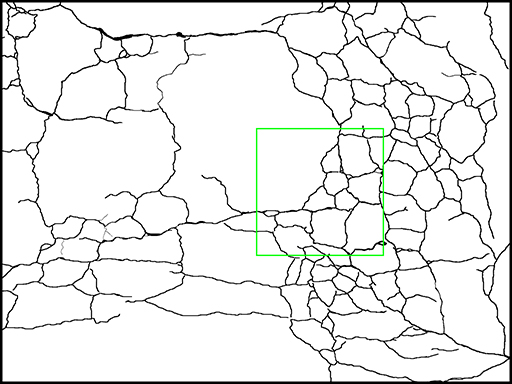}%
\label{7-2-4}}
\vspace{-5pt}

\subfloat{\includegraphics[width=1.65in]{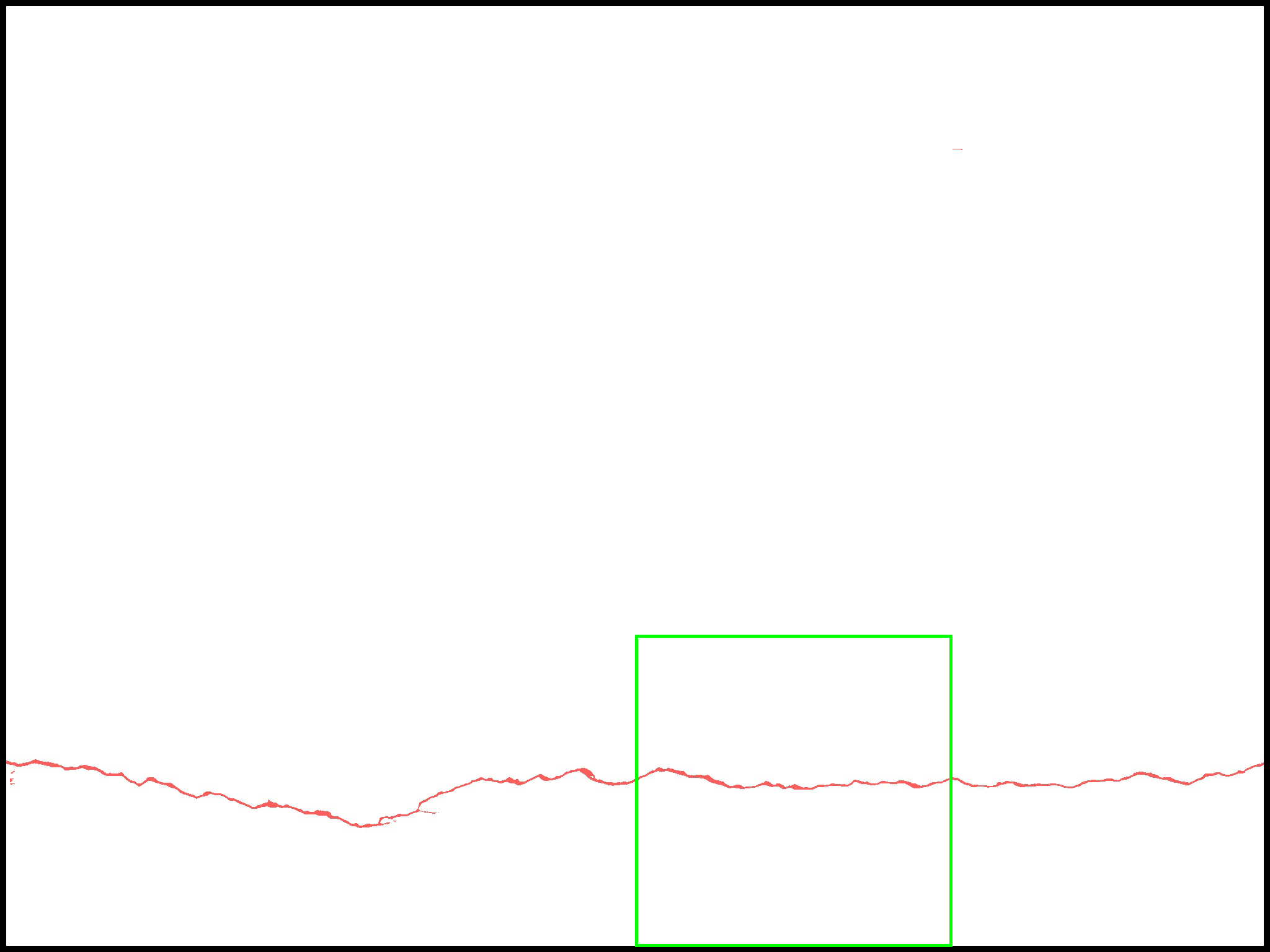}%
\label{7-3-1}}
\hfil
\subfloat{\includegraphics[width=1.65in]{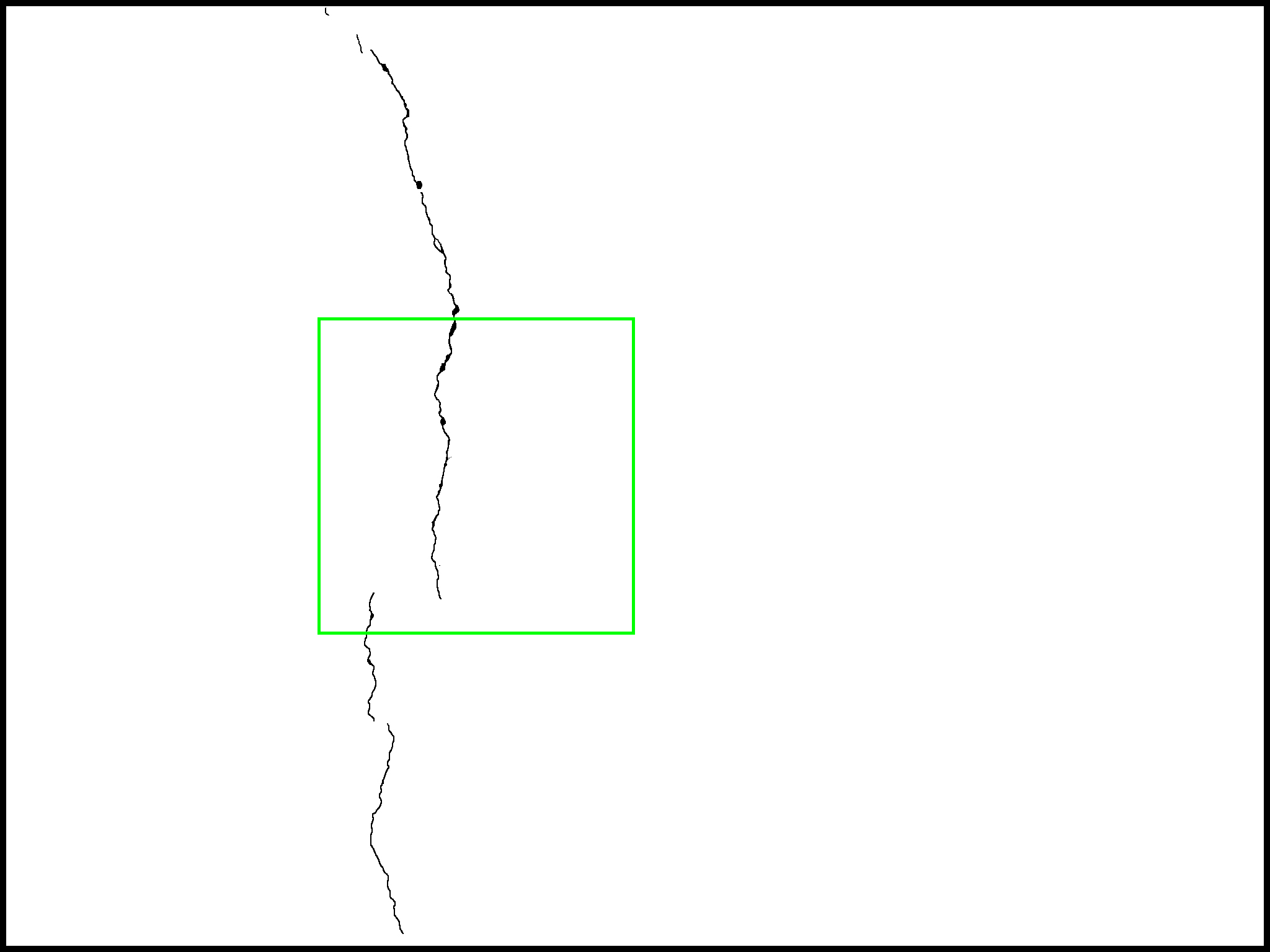}%
\label{7-3-2}}
\hfil
\subfloat{\includegraphics[width=1.65in]{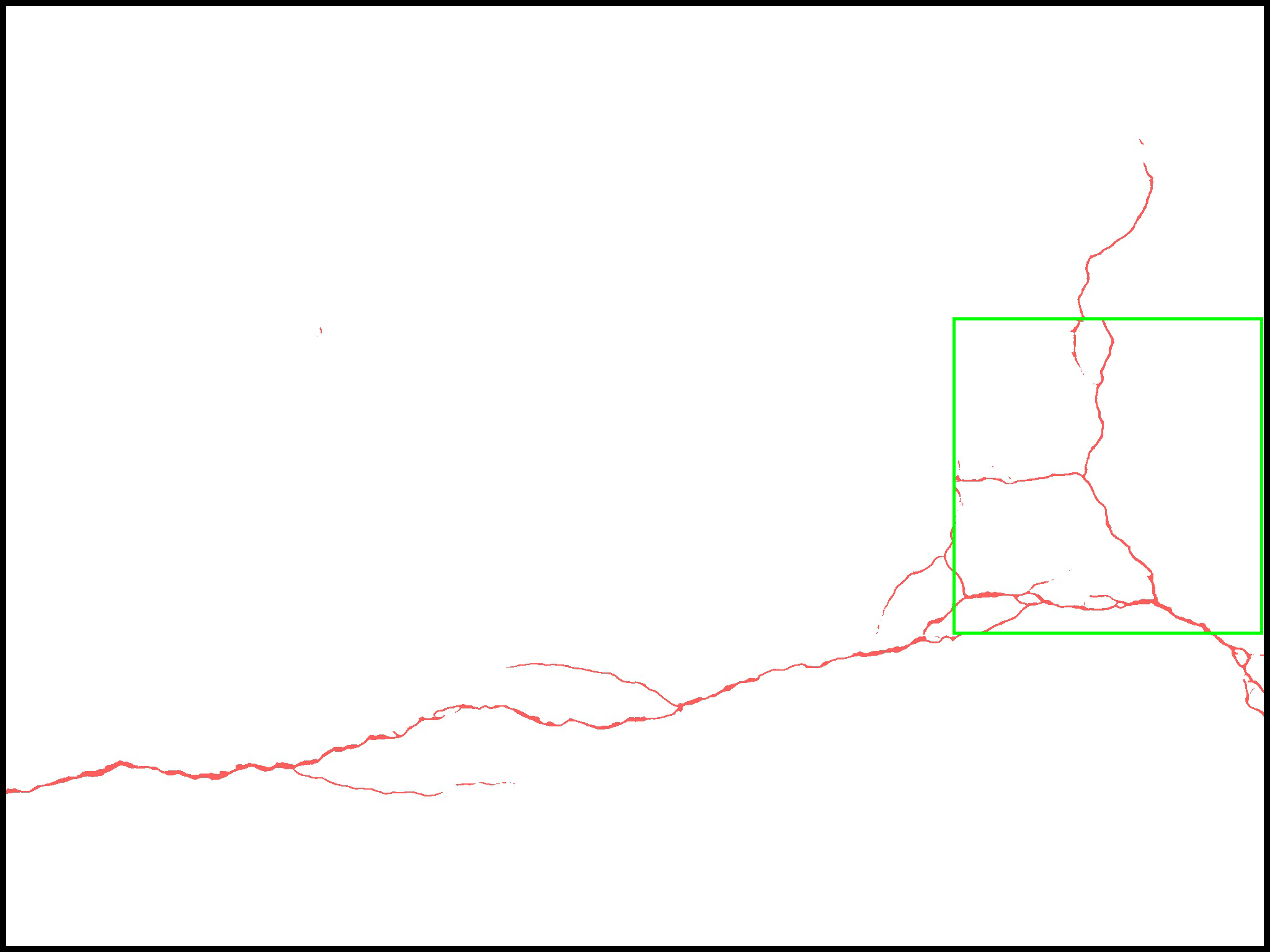}%
\label{7-3-3}}
\hfil
\subfloat{\includegraphics[width=1.65in]{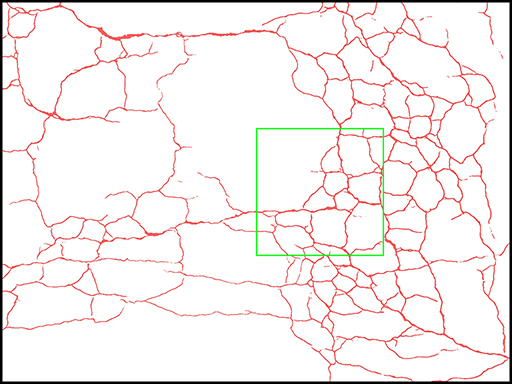}%
\label{7-3-4}}
\vspace{-5pt}

\subfloat{\includegraphics[width=1.65in]{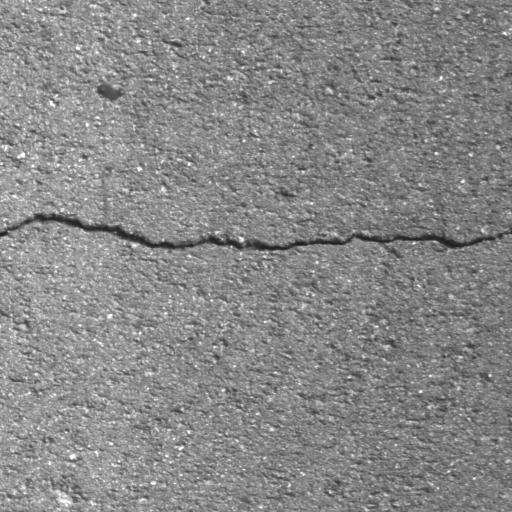}%
\label{7-4-1}}
\hfil
\subfloat{\includegraphics[width=1.65in]{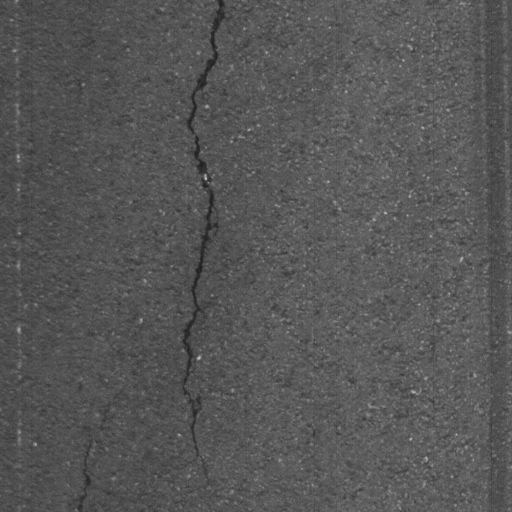}%
\label{7-4-2}}
\hfil
\subfloat{\includegraphics[width=1.65in]{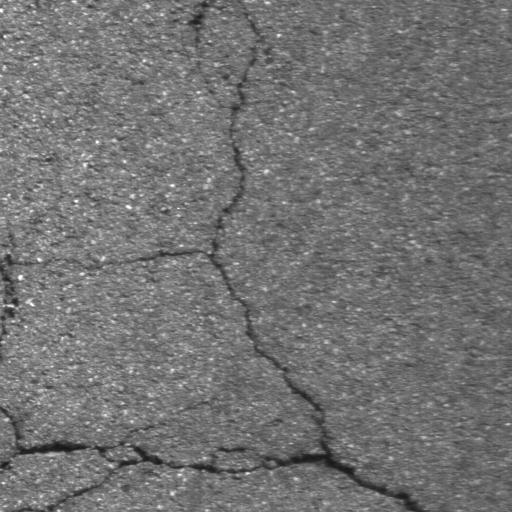}%
\label{7-4-3}}
\hfil
\subfloat{\includegraphics[width=1.65in]{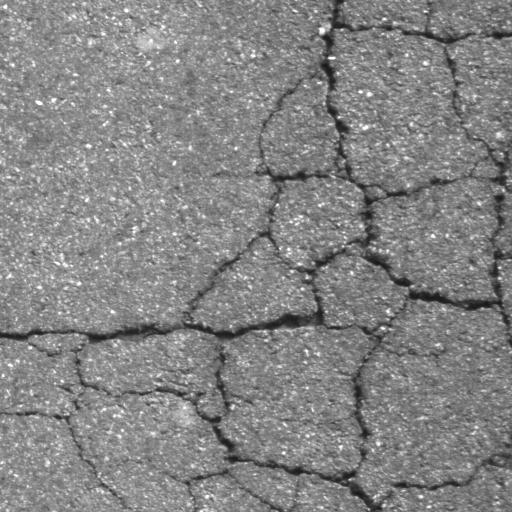}%
\label{7-4-4}}
\vspace{-5pt}

\subfloat{\includegraphics[width=1.65in]{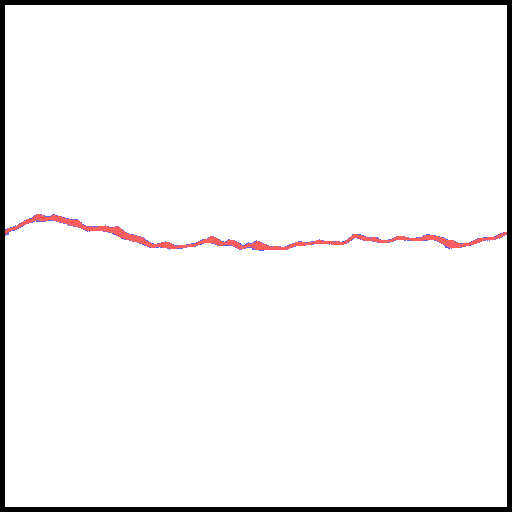}%
\label{7-5-1}}
\hfil
\subfloat{\includegraphics[width=1.65in]{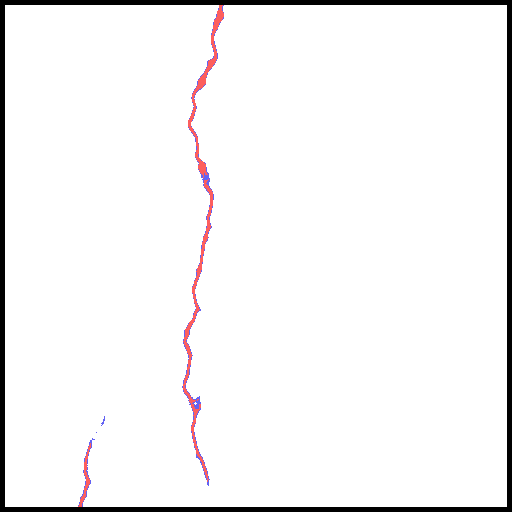}%
\label{7-5-2}}
\hfil
\subfloat{\includegraphics[width=1.65in]{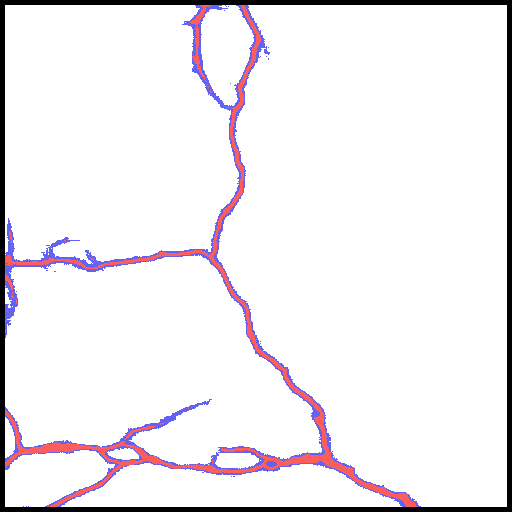}%
\label{7-5-3}}
\hfil
\subfloat{\includegraphics[width=1.65in]{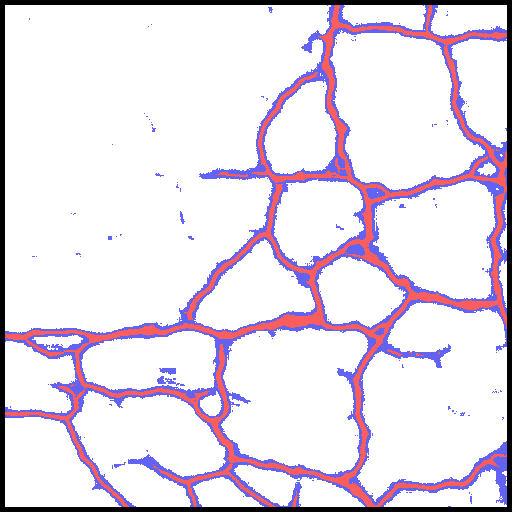}%
\label{7-5-4}}
\vspace{-5pt}

\subfloat{\includegraphics[width=1.65in]{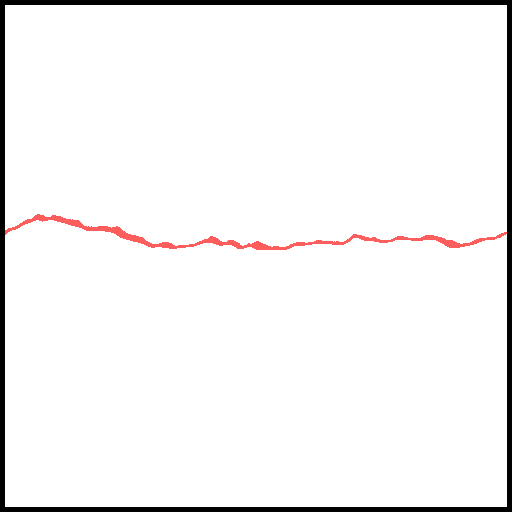}%
\label{7-6-1}}
\hfil
\subfloat{\includegraphics[width=1.65in]{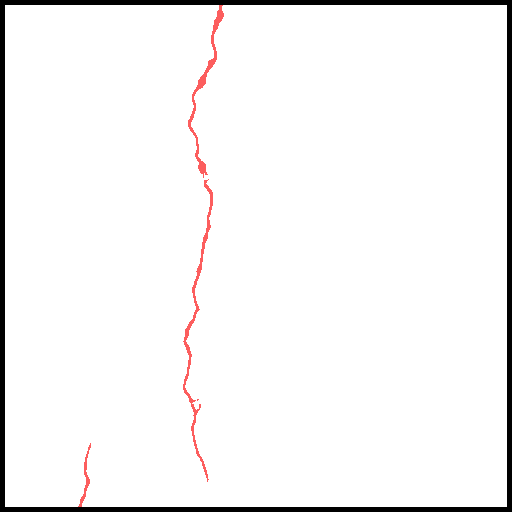}%
\label{7-6-2}}
\hfil
\subfloat{\includegraphics[width=1.65in]{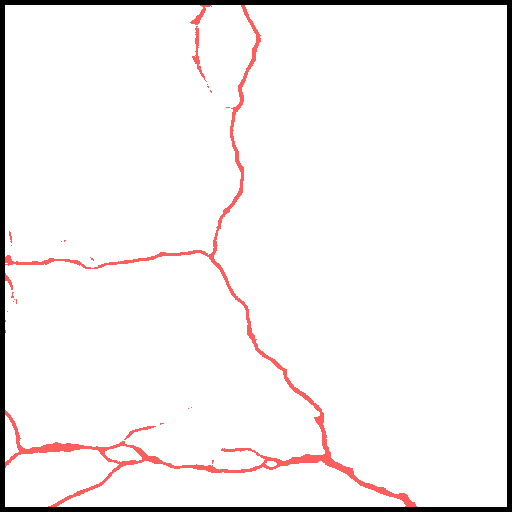}%
\label{7-6-3}}
\hfil
\subfloat{\includegraphics[width=1.65in]{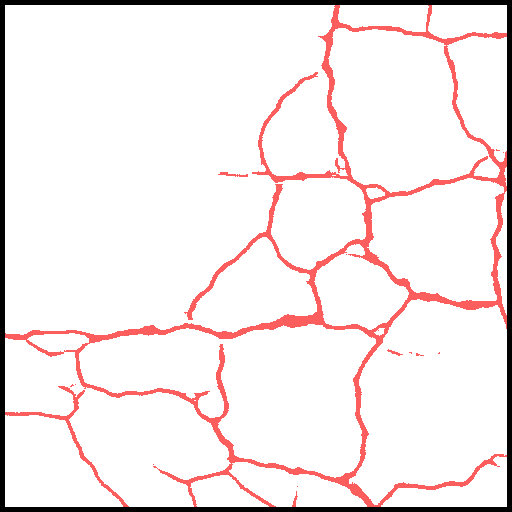}%
\label{7-6-4}}
\hfil

\centering
\caption{Typical results of four types of cracks on G45 dataset using FPCNet (from left to right: transverse cracks, longitudinal cracks, block cracks, alligator cracks; from top to bottom: original image, ground truth, predicted image, detailed images having green rectangles in first row, detailed images of probability image, detailed images of predicted image).}
\label{fig_7}
\end{figure*}

The G45 crack dataset consists of 122 grayscale crack images with a resolution of $2048 \times 1536$; these images were collected on the G45 highway in China. The ground truth of each image is carefully labeled by professional engineers. The dataset covers four types of cracks: transverse cracks, longitudinal cracks, block cracks, and alligator cracks. The imaging conditions of each image in the dataset are not uniform and the imaging brightness is different, resulting in a difference in contrasts between the crack and the pavement in the image. Moreover, the number, length and width of the cracks in the image vary. Some images contain oil stains, speckle noise, and lane lines, which makes crack detection difficult.

We randomly divided the 122 crack images into 77 training images and 45 testing images. Owing to the large image resolution, each image is cropped into twelve images having a size of $512 \times 512$ (without overlapping), and then these images are sent to the network for training.

\subsubsection{Implementation details}The data augmentation is performed in the same manner as on the CFD dataset except for the random color jittering because the images in the G45 dataset are in grayscale. Further, we also omit the clockwise rotation of $180^{\circ}$. Random cropping is used in training, but with a size of $480 \times 480$. We use the same hyperparameters as those when training on the CFD dataset, except for the batch size of 2.

\subsubsection{Results}To obtain the complete prediction result, we crop the entire $2048 \times 1536$ pavement image into twelve $512 \times 512$ patches and send them to the network separately. Next, we combine the twelve predicted results into an overall prediction of the entire pavement image.

Fig. \ref{fig_7} shows the typical results of the four types of cracks. The following aspects are shown from top to bottom: pavement crack images, ground truth, prediction results from FPCNet, detailed images having green rectangles in the first row, and probability results of detailed images and prediction results of detailed images. From left to right, the results for transverse cracks, longitudinal cracks, block cracks, and alligator cracks, are shown. FPCNet can analyze different types of cracks with sufficient contextual features of the MD module. As a result, all cracks can be robustly detected irrespective of the complex pattern they possess (third row in Fig. \ref{fig_7}).

FPCNet demonstrates excellent ability to identify transverse cracks (first column in Fig. \ref{fig_7}) owing to its simple structure and the effect of the network. Only little noise exists (blue pixels) in the probability image shown in the fifth row, which indicates that the proposed network detects transverse cracks with a remarkably high accuracy. Because the data augmentation of rotation by $90^{\circ}$ is used during the training, the rotated images of the transverse cracks enhance the recognition ability of the longitudinal cracks. As shown in the second column of Fig. \ref{fig_7}, longitudinal cracks are detected with high accuracy, similar to the transverse cracks. Block cracks are more complicated than transverse and longitudinal cracks (third column in Fig. \ref{fig_7}). Missed detections occur in certain location in which the cracks are shallow and indistinct. As seen in the fifth row, FPCNet assigns them a low prediction probability. However, the proposed network can still detect most cracks of this type reasonably well (final row in third column). Alligator cracks are the most complex type of pavement cracks (fourth column in Fig. \ref{fig_7}) and they are densely interlaced. As illustrated in fifth row of the fourth column, more noise exists in this probability image compared to those of other types of cracks; however, after binarization, the images with low probability are filtered. This proves that the proposed network demonstrates satisfactory performance for the detection of alligator cracks.

\begin{figure}[p]
\centering

\subfloat{\includegraphics[width=1.65in]{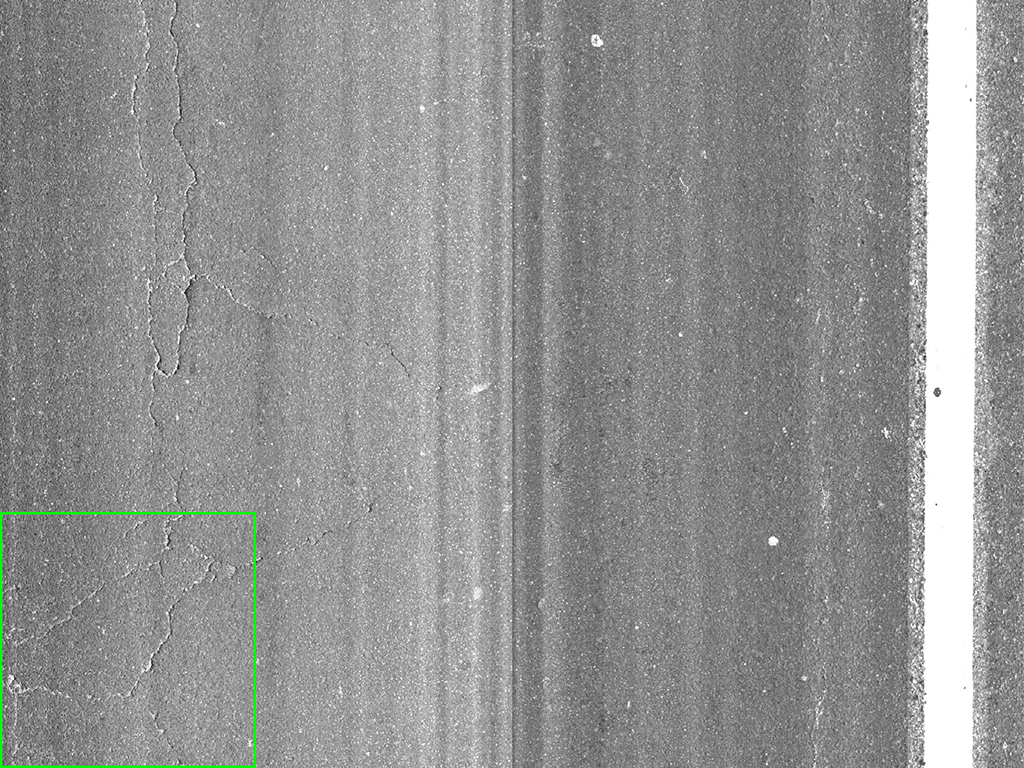}%
\label{8-1-1}}
\hfil
\subfloat{\includegraphics[width=1.65in]{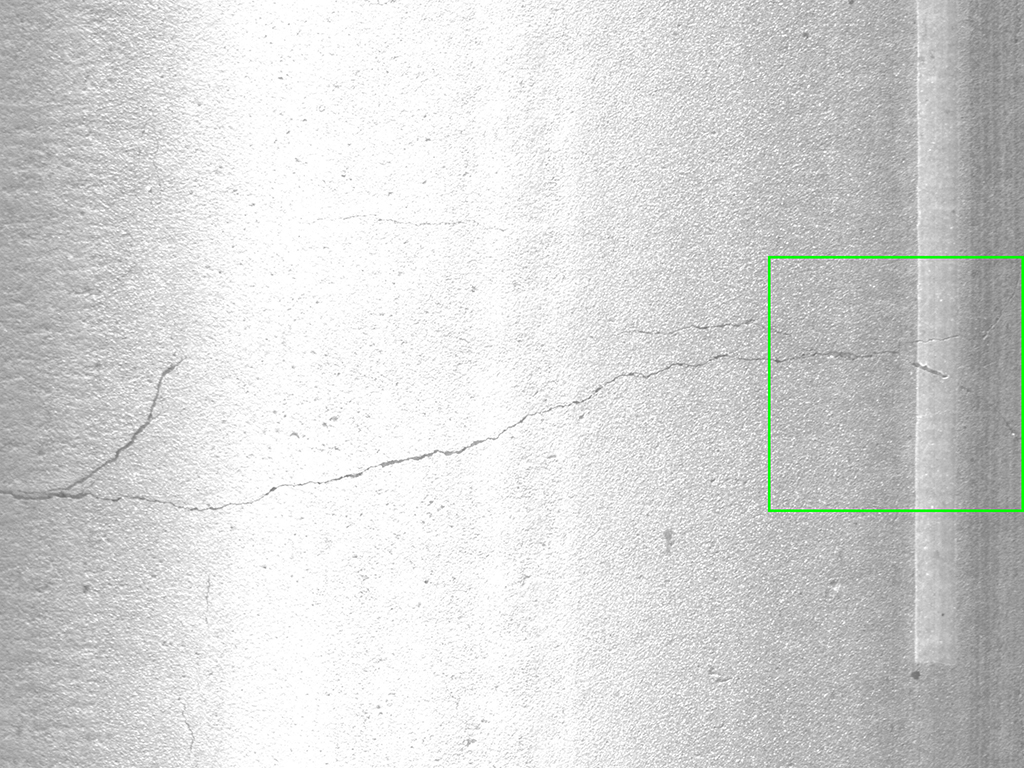}%
\label{8-1-2}}
\vspace{-5pt}

\subfloat{\includegraphics[width=1.65in]{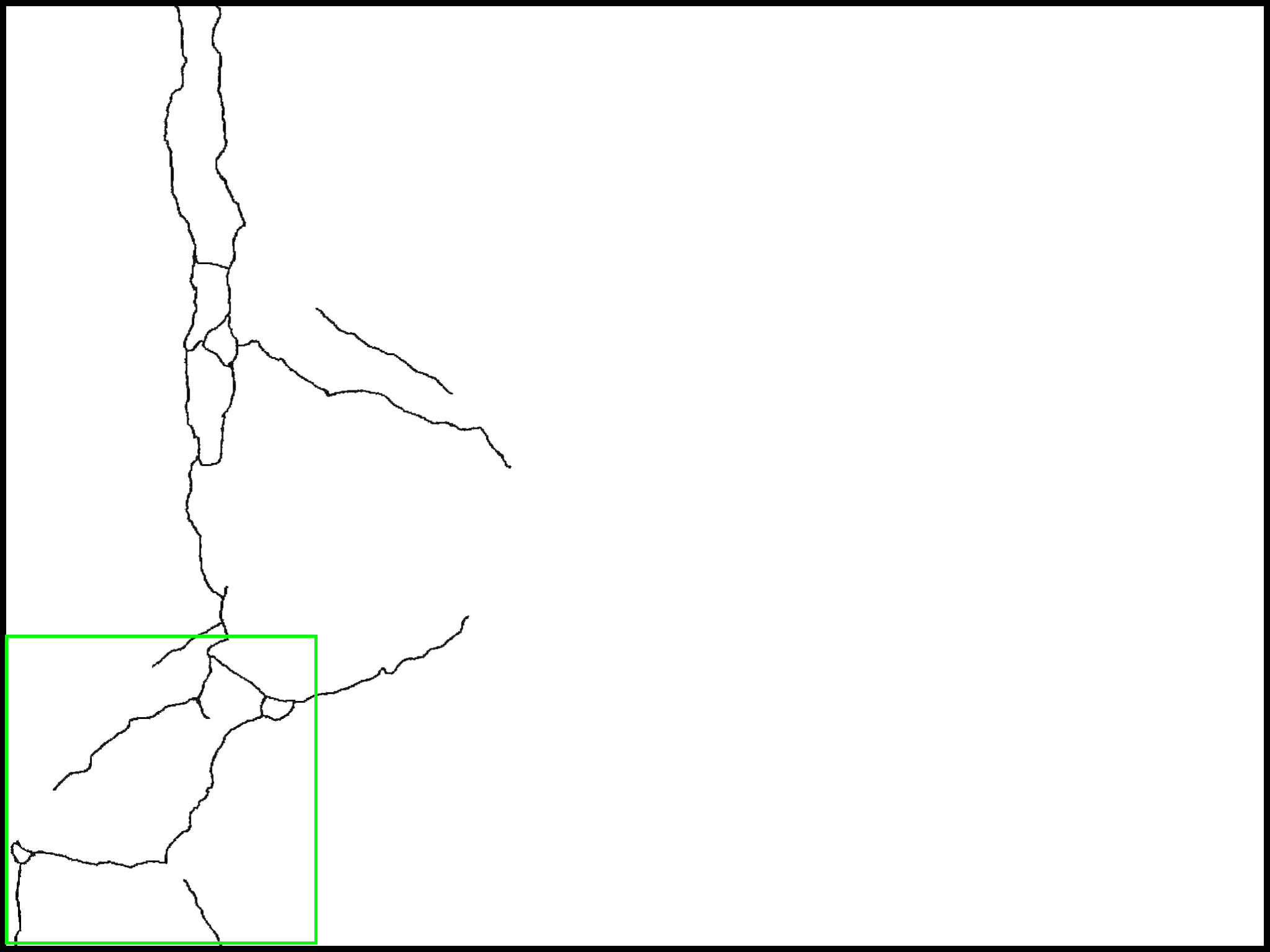}%
\label{8-2-1}}
\hfil
\subfloat{\includegraphics[width=1.65in]{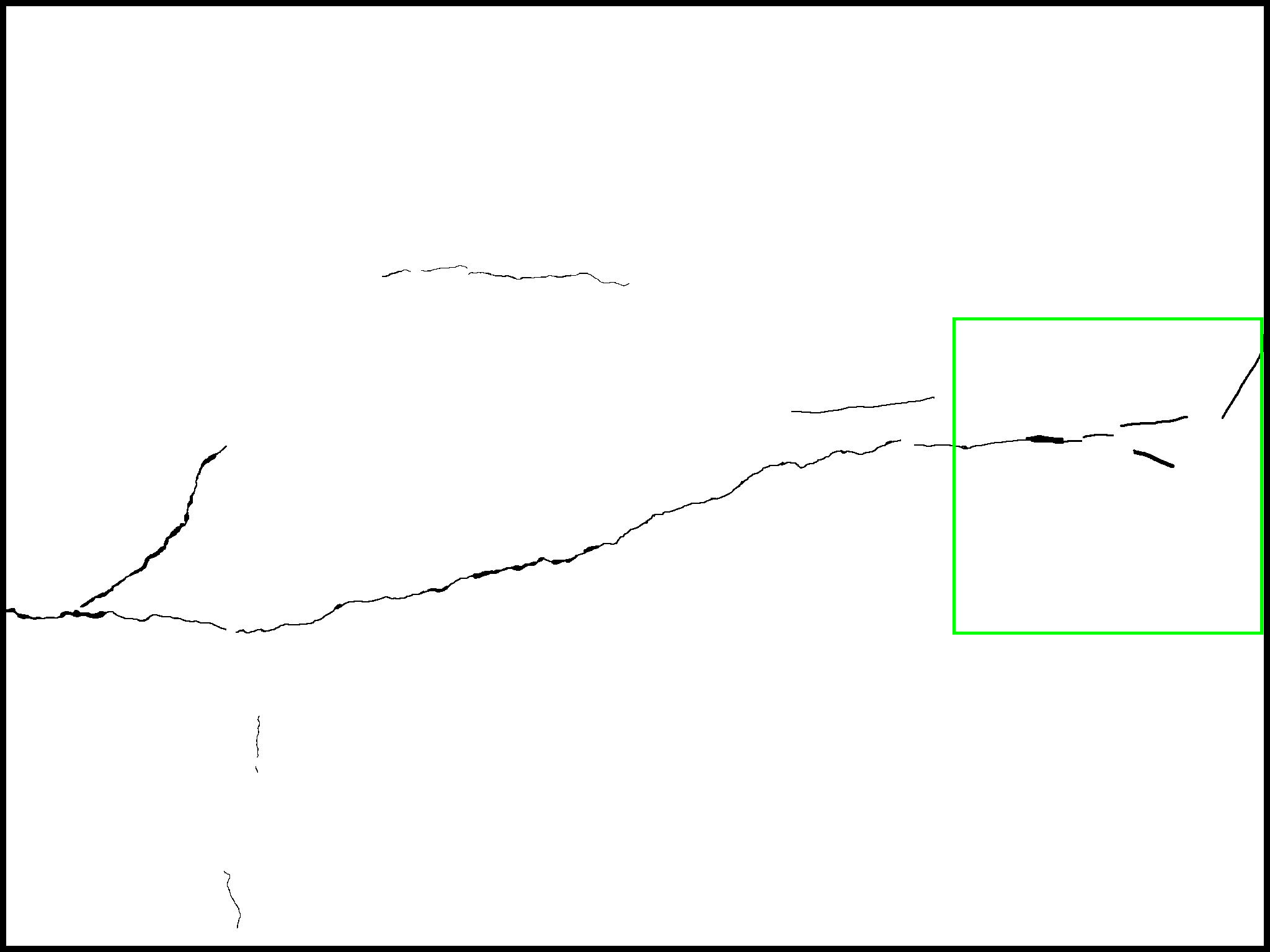}%
\label{8-2-2}}
\vspace{-5pt}

\subfloat{\includegraphics[width=1.65in]{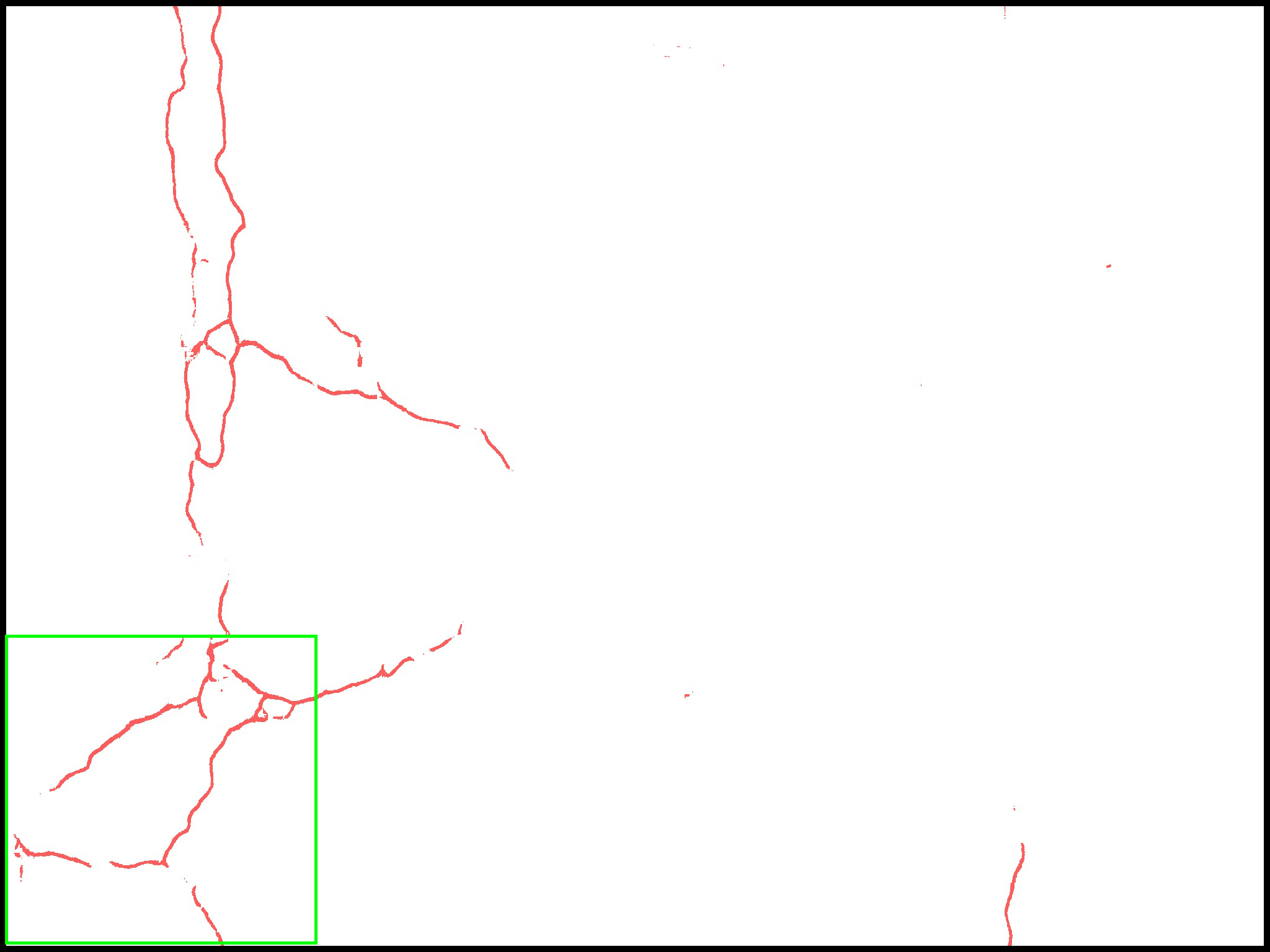}%
\label{8-3-1}}
\hfil
\subfloat{\includegraphics[width=1.65in]{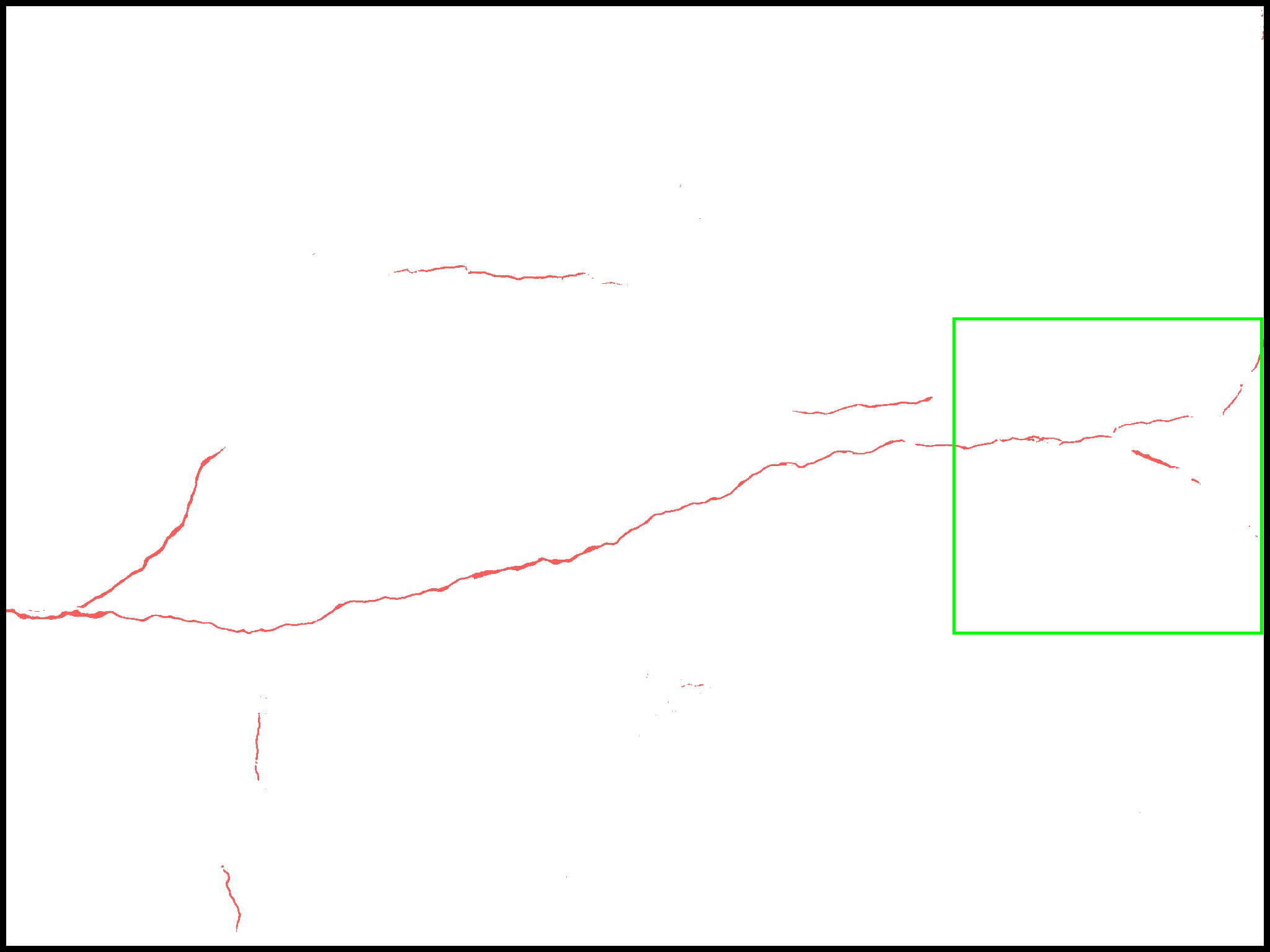}%
\label{8-3-2}}
\vspace{-5pt}

\subfloat{\includegraphics[width=1.65in]{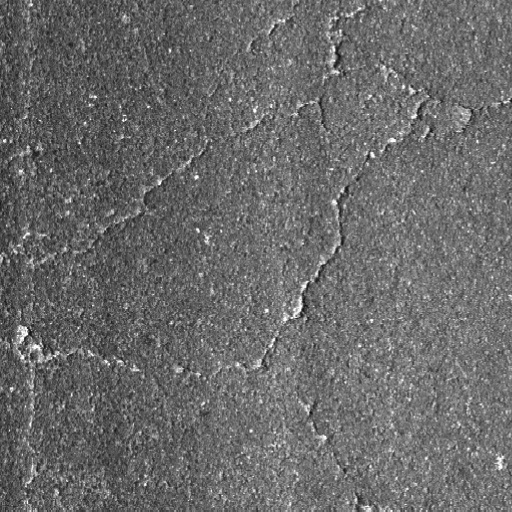}%
\label{8-4-1}}
\hfil
\subfloat{\includegraphics[width=1.65in]{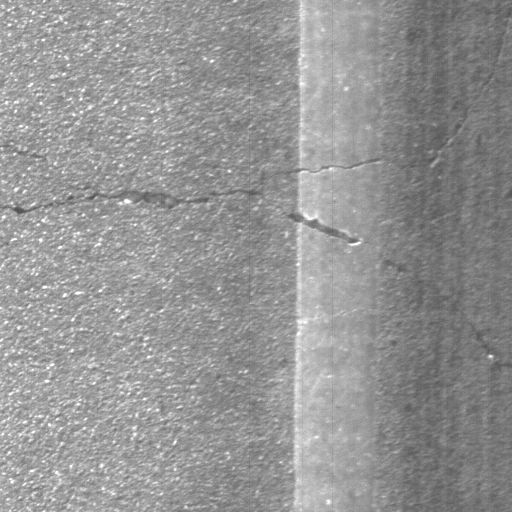}%
\label{8-4-2}}
\vspace{-5pt}

\subfloat{\includegraphics[width=1.65in]{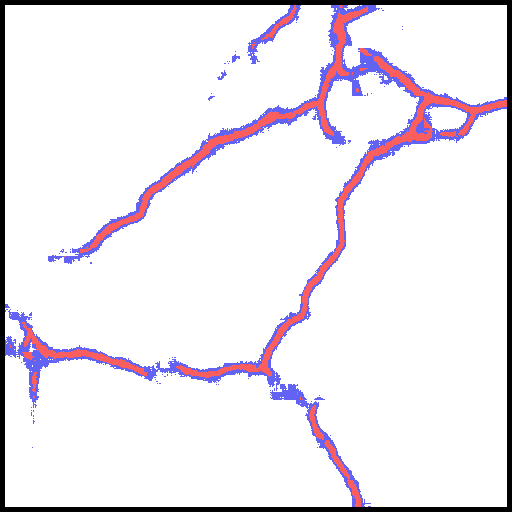}%
\label{8-5-1}}
\hfil
\subfloat{\includegraphics[width=1.65in]{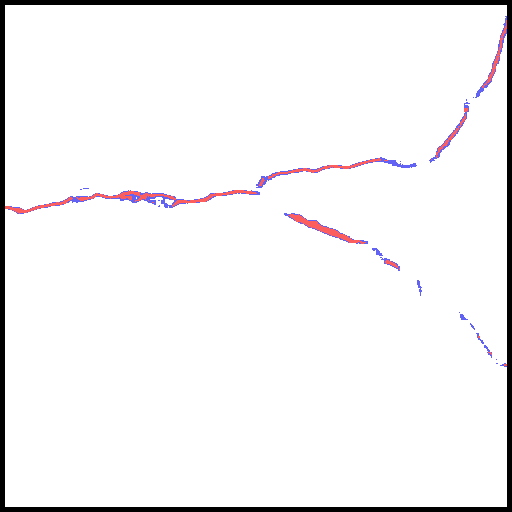}%
\label{8-5-2}}
\vspace{-5pt}

\subfloat{\includegraphics[width=1.65in]{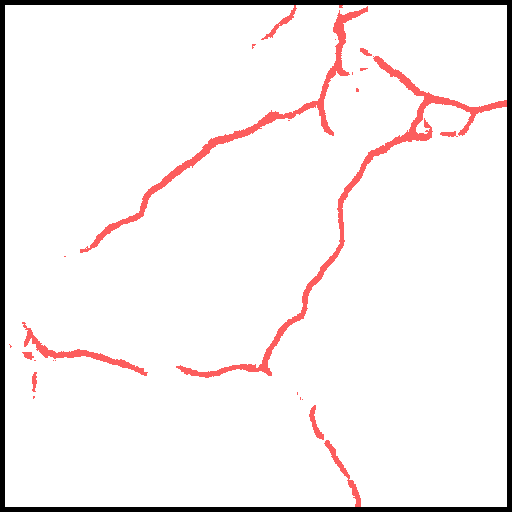}%
\label{8-6-1}}
\hfil
\subfloat{\includegraphics[width=1.65in]{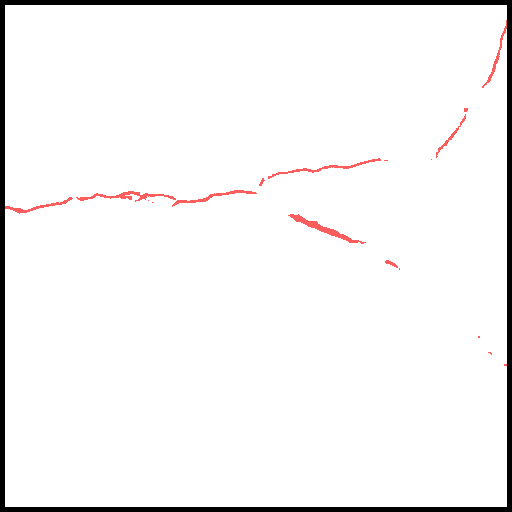}%
\label{8-6-2}}
\hfil

\centering
\caption{Typical examples of crack images with low contrast, zebra crossings, and noise.}
\label{fig_8}
\end{figure}

We also illustrate some typical examples of crack images with low contrast, zebra crossings, and noise in Fig. \ref{fig_8}. As seen in this figure, such cracks are more indiscernible than other cracks and their backgrounds are full of noise. In addition, the image in the right column also exhibits poor illumination. The presence of such features in crack images critically affects the detection of cracks, leading to an increase in missed detections. However, most cracks can be successfully detected by FPCNet, indicating the robustness of the proposed network.

The evaluation results of the four types of cracks on the G45 dataset are presented in TABLE \ref{table_3}. Transverse and longitudinal cracks have relatively good detection results owing to their simple structure and our employed effective training strategy mentioned above. The F1 scores of both these types of cracks reach values more than 95\%. Most images of block cracks images involve bad imaging conditions and noise, leading to an increase in missed detection. However, owing to the robustness of FPCNet, the Precision, Recall and F1 score of block cracks still reach values of 91.53\%, 89.90\%, and 90.71\%, respectively. Our network performs robustly in the case of alligator cracks, for which the values of Precision, Recall and F1 score are 95.24\%, 94.26\%, and 94.75\%, respectively.

\begin{table}[!t]
\renewcommand{\arraystretch}{1.5}
\caption{EVALUATION RESULTS FOR DIFFERENT TYPES OF CRACKS ON G45 DATASET}
\label{table_3}
\centering
\begin{tabular}{cccc}
\hline
Type & Precision & Recall & F1 score\\
\hline
Transverse & 98.22\% & 96.82\% & 97.51\%\\
Longitudinal & 93.10\% & 98.58\% & 95.76\%\\
Block & 91.53\% & 89.90\% & 90.71\%\\
Alligator & 95.24\% & 94.26\% & 94.75\%\\

\hline
\end{tabular}
\end{table}

\begin{table}[!t]
\renewcommand{\arraystretch}{1.5}
\caption{EVALUATION RESULTS FOR DIFFERENT DILATION RATES ON CFD DATASET}
\label{table_4}
\centering
\begin{tabular}{cccc}
\hline
Dilation rates & Precision & Recall & F1 score\\
\hline
\{1, 2, 3, 4\} & \textbf{97.48\%} & \textbf{96.39\%} & \textbf{96.93\%}\\
\{1, 2, 4, 8\} & 97.00\% & 96.29\% & 96.64\%\\
\{2, 4, 8, 16\} & 96.92\% & 96.36\% & 96.64\%\\
\hline
\end{tabular}
\end{table}

\begin{table}[!t]
\renewcommand{\arraystretch}{1.5}
\caption{EVALUATION RESULTS FOR DIFFERENT DILATION RATES ON G45 DATASET}
\label{table_5}
\centering
\begin{tabular}{cccc}
\hline
Dilation rates & Precision & Recall & F1 score\\
\hline
\{1,2,3,4\} & \textbf{95.01\%} & \textbf{93.94\%} & \textbf{94.47\%}\\
\{1,2,4,8\} & 94.96\% & 93.47\% & 94.21\%\\
\{2,4,8,16\} & 94.84\% & 93.50\% & 94.16\%\\
\hline
\end{tabular}
\end{table}

\subsection{Discussion}

The dilation rate introduced in Eqn. \ref{equa_1} is an important hyperparameter that allows us to vary the context size obtained by the MD module in the network. As illustrated in Fig. \ref{fig_1}, a larger dilation rate corresponds to a larger context size that can be obtained. Different context sizes can lead to different effects on the prediction results. To investigate the appropriate combination of different dilation rates for crack detection, we conduct experiments on the CFD and G45 datasets to discuss the setting of the hyperparameters in the MD module.

Three dilation rates groups of \{1, 2, 3, 4\}, \{1, 2, 4, 8\}, \{2, 4, 8, 16\} are tested. As seen from the results given in TABLE \ref{table_4} and TABLE \ref{table_5}, the highest accuracy is achieved on both datasets when the dilation rates are of the group of \{1, 2, 3, 4\}. This is because for a relatively elongated structure such as a crack, a larger dilation rate ignores more details of the cracks, thereby causing a decrease in the accuracy.

\section{Conclusion}
In this paper, we propose a high-precision and high-speed pavement crack detection network called FPCNet. A Multi-Dilation module and an SE-Upsampling module are developed in this framework. The Multi-Dilation module extracts the crack MD features of multiple context sizes to robustly detect cracks with different widths and topologies. The SE-Upsampling module restores the resolution of the MD features and assigns different weights to the MD features after lateral connection, to optimize the MD features. By integrating these two modules in the Encoder-Decoder structure, FPCNet characterizes the crack context with the MD module, and recursively optimizes the contextual features step-by-step. Finally, pixel-level crack detection is achieved.

The results of a large number of experiments performed on two different crack datasets prove that the proposed method can robustly detect many types of cracks and achieve state-of-the-art results on the CFD dataset, with a speed of 14.7 FPS.

In future work, we will test our FPCNet in more crack datasets. In addition, learning-based conditional random field (CRF) will be performed in the network to further refine its output.


%



\ifCLASSOPTIONcaptionsoff
  \newpage
\fi



\bibliographystyle{IEEEtran}
%

\bibliography{references}



%





\end{document}